\documentclass[lettersize,journal]{IEEEtran}
\usepackage{amsmath,amsfonts}
\usepackage{algorithmic}
\usepackage[ruled,vlined]{algorithm2e}
\usepackage{amssymb}
\usepackage{array}
\usepackage[caption=false,font=footnotesize,labelfont=sf,textfont=sf]{subfig}
\usepackage{textcomp}
\usepackage{stfloats}
\usepackage{url}
\usepackage{verbatim}
\usepackage{graphicx}
\usepackage{xcolor}
\usepackage{booktabs}
\usepackage{multirow}
\usepackage[font = footnotesize]{caption}
\usepackage{makecell} 
\usepackage[table]{xcolor} %
\usepackage{soul, color, xcolor}
\usepackage{hyperref}

\newtheorem{theorem}{Theorem}[section]
\newtheorem{Problem}[theorem]{Problem}

\definecolor{green}{HTML}{6aa84f} %
\definecolor{highlight}{HTML}{cfe2f3} %
\definecolor{blue}{HTML}{0000FF} %
\definecolor{brown}{HTML}{9F8C76}
\definecolor{pink}{HTML}{D88782}
\definecolor{light-blue}{HTML}{4a86e8}
\definecolor{dark-yellow}{HTML}{F6AE2D}
\definecolor{light-pink}{HTML}{ffcccc}
\definecolor{light-yellow}{HTML}{FFE5B4}

\newcommand{\brown}[1]{{\color{brown}#1}}
\newcolumntype{L}[1]{>{\raggedright\arraybackslash}p{#1}}
\newcolumntype{C}[1]{>{\centering\arraybackslash}p{#1}}
\AtBeginEnvironment{table}{\footnotesize}

\newcommand{\ours}{DR-LfD\xspace}

\newcommand{\pddl}[1]{\texttt{#1}}
\newcommand{\pddlkw}[1]{\textbf{\pddl{#1}}}
\usepackage{listings}
\def\BibTeX{{\rm B\kern-.05em{\sc i\kern-.025em b}\kern-.08em
    T\kern-.1667em\lower.7ex\hbox{E}\kern-.125emX}}
\usepackage{balance}

\begin{document}

\makeatletter
\let\@oldmaketitle\@maketitle
\renewcommand{\@maketitle}{\@oldmaketitle%
  \begin{center}
  \vspace{0.2in}
    \setcounter{figure}{0}
    \includegraphics[width=1.0\textwidth]{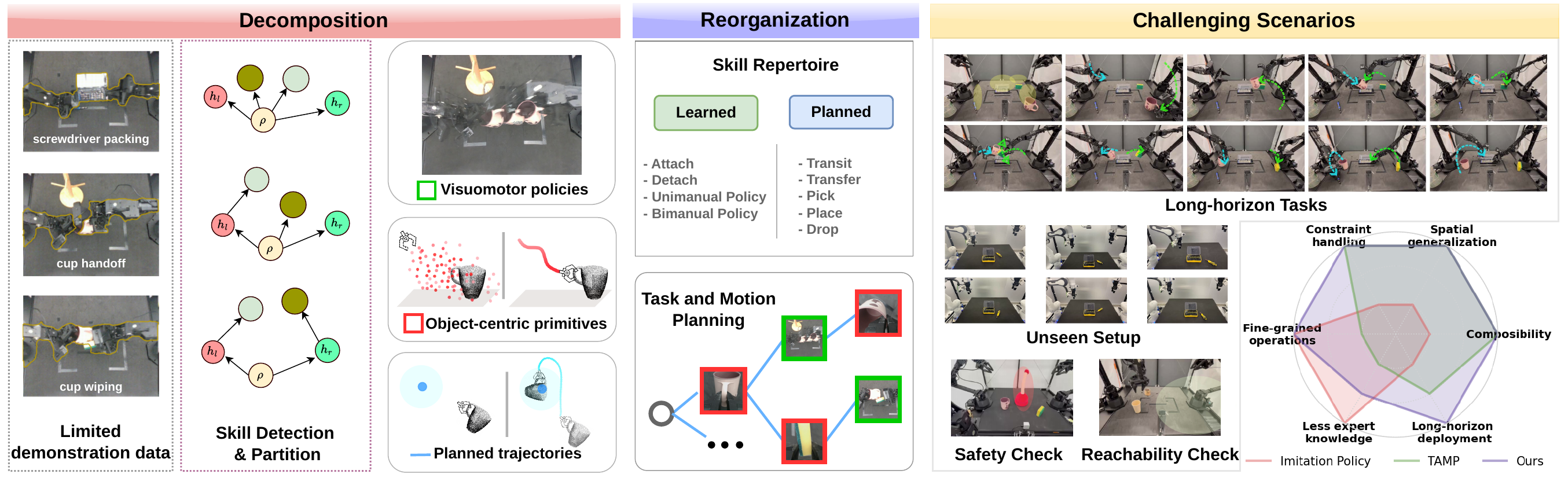}
   \vspace{-0.2in}
    \captionof{figure}{
     We present \ours, a systematic framework that decomposes demonstrations into skills and reorganizes them for novel tasks.
    \ours adopts a graph-like representation of demonstrated skills, segmenting and categorizing skills according to changes in contact relationships. The symbolic skill repertoire wraps visuomotor policies, object-centric primitives, and planned motions, all of which are used to compose plans by TAMP. \ours combines the strength of imitation policies and TAMP: with a small amount of demonstrations for each task, \ours exhibits strong performance on tasks with multiple steps, unseen setups, and physical constraints.}
    \label{fig:teaser}
    \vspace{-0.4in}
  \end{center}
}
\makeatother

\title{
Decompose and Reorganize: Planning with  Primitives and Visuomotor Policies Learned from Demonstrations
}
\author{
    Yizhou~Chen, Hang~Xu, Dongjie~Yu, Yupu~Lu, Tengye~Xu, Zeqing~Zhang, Wei~Zhang, Yi~Ren, Ben~M.~Chen, and~Jia~Pan\\
    {\normalsize\hypersetup{pdfborder={0 0 0}}\href{https://dr-lfd.github.io/DR-LfD-website}{\textcolor{light-blue}{\nolinkurl{https://dr-lfd.github.io/DR-LfD-website}}}}%
\thanks{Yizhou Chen, Dongjie Yu, Yupu Lu, Tengye Xu, and Jia Pan are with The University of Hong Kong, Hong Kong.}%
\thanks{Hang Xu is with JD.com (JingDong).}%
\thanks{Zeqing Zhang is with Nanyang Technological University (NTU), Singapore.}%
\thanks{Wei Zhang is with Southern University of Science and Technology (SUSTech), Shenzhen, China.}%
\thanks{Yi Ren is with Huawei Technologies.}%
\thanks{Ben M. Chen is with The Chinese University of Hong Kong (CUHK), Hong Kong.}%
}

\maketitle

\begin{abstract}
Successfully automating dexterous, long-horizon bimanual manipulation requires frameworks capable of both high-level reasoning and fine-grained execution.  Traditional task and motion planning (TAMP), while excellent at symbolic planning, is often brittle in contact-rich operations. Simultaneously, imitation learning (IL), while effective in manipulation tasks with visual feedback, is limited by its low capability in spatial generalization and multi-stage operation.   To reconcile their complementary strengths and limitations, we propose \textit{\ours} (\underline{D}ecomposed and \underline{R}eorganized Skills \underline{L}earned \underline{f}rom \underline{D}emonstrations), a framework that integrates visuomotor policies into a TAMP-gated decision-making system.
Based on contact relationships, \ours  decomposes human demonstrations into atomic skills, which are reproduced as visuomotor policies or object-centric primitives.
The initiation, termination, and constraints of the visuomotor policies are carefully modeled and implemented in a TAMP-compatible form, enabling reorganization of skills learned from different sources. 
\ours transforms the learning problem from one requiring exponential demonstration data over possible skill sequences to one whose demonstration burden scales with the number of distinct skill types, with limited data for each skill.
Real-world and simulation benchmarks demonstrate data-efficient skill composition under unseen setups and physical constraints. On two DexMimicGen tasks, \ours with 100 demonstrations outperforms diffusion-based imitation policies with 1000.
\end{abstract}


\section{Introduction}

Task and motion planning (TAMP) provides a powerful framework for long-horizon robotic manipulation by leveraging symbolic reasoning to sequence discrete skills and their continuous motions. However, its practical application is often hindered by dependencies on manually engineered models, perfect perception, and flawless execution. For instance, while methods like IDTMP \cite{dantam2018incremental} and TAMPER \cite{pan2024task} incorporate feedback and replanning, they still rely on precise object pose estimation, making them brittle in contact-rich, unstructured environments.

In parallel, Learning-from-demonstration (LfD) offers an alternative paradigm, bypassing the need for explicit modeling by learning visuomotor policies directly from data. With the help of accessible teleoperation devices \cite{zhao2023learning,wu2024gello}, advanced imitation learning  (IL) methods \cite{chi2023diffusion, florence2022implicit,mandlekar2021matters,ke20243d,ze20243d} stand out in effectively replicating complex, multi-modal behaviors.
Nonetheless, these end-to-end policies remain highly vulnerable to out-of-distribution (OOD) observations. A slight displacement of an object or an unseen grasp pose can lead to failure, and errors compound over long skill sequences. To combat this, recent works \cite{black2024pi0}, \cite{intelligence2025pi05}, \cite{zhao2024aloha, liu2024rdt} pursue large-scale pre-training of behavior models \cite{barreiros2025careful}, but the high costs in data and computation hardware are a significant barrier for applications needing rapid, lightweight deployment.

Learning long-horizon tasks is further complicated by the many possible combinations of individual skills. Consider a robot with a repertoire of $N$ manipulation skills. To accomplish a novel task requiring a sequence of $M$ such skills, an end-to-end policy must implicitly learn not only each skill but also the decision-making for when, which, and how to switch to the next skill -- e.g., determining the skill progress, transition timing, and whether a motion plan is needed or if a regrasping operation is necessary. This requires the policy to internalize a combinatorial number of possible sequences, demanding exponentially large datasets for effective generalization. 
Furthermore, the diversity of the demonstration data is constrained by both the workspace limitations of the teleoperation hardware \cite{zhao2023learning, wu2024gello} and the inherent kinematic limitations of the human operator, exacerbating the problem. 

TAMP and LfD are complementary, yet naively combining them leaves three questions open. How can a planned motion hand over to a learned skill at a state the skill can handle? How can TAMP reason over learned and planned skills in a uniform way? And how can TAMP verify that a learned skill is feasible before executing it?
To answer these questions, we propose \textit{\ours} (\underline{D}ecomposed and \underline{R}eorganized Skills \underline{L}earned \underline{f}rom \underline{D}emonstrations), a framework that integrates the strength of both paradigms to achieve data-efficient and reliable long-horizon manipulation.
The core idea is to decompose the problem: we learn individual manipulation skills from a small number of demonstrations and reorganize them using a TAMP solver, which connects them through planned trajectories and geometric reasoning to achieve combinatorial goals.
As a result, the demonstration requirement scales with the number of distinct skill types, rather than exponentially with the length of skill sequences.
{Beyond data efficiency, this decomposition lets an individual skill be improved in isolation for long-horizon operations, without recollecting data for the entire task.}

{Fig. \ref{fig:teaser} illustrates this pipeline: \ours segments demonstrations into atomic skills and wraps each skill as a symbolic operator for the TAMP solver. This TAMP-gated system uses a skill repertoire with learned visuomotor policies, learned object-centric primitives, and planned motions to achieve synergistic effects. }

The contributions of this work are fourfold:
\begin{itemize}
    \item We devise a novel TAMP-gated system that abstracts atomic skills from demonstrations and reorganizes them to achieve combinatorial goals;
    \item We leverage equivariant diffusion models to train object-centric primitives, enabling robust spatial generalization from minimal data;
    \item We systematically model the initiation, termination, and constraints of learned visuomotor policies in a form compatible with a sampling-based TAMP framework;
    \item We demonstrate through comprehensive real-world and simulated experiments that \ours completes long-horizon manipulation tasks under challenging conditions while complying with physical constraints, significantly outperforming state-of-the-art imitation learning methods.
\end{itemize}

\section{Related work}  \label{sec:relatedwork}%

\subsection{Planning and Imitation for Bimanual Tasks} %

We set our focus on bimanual manipulation, where two robot arms work jointly to achieve challenging and human-like maneuvers \cite{krebs2022bimanual,yu2024bikc, chen2019combined}. Traditional planning approaches often fall short in this area due to requirements of precise hand-hand and hand-eye coordination \cite{ren2024enabling}. There are traditional TAMP methods dedicated to dual-arm robot systems \cite{takata2022efficient, chen2019combined}. However, these approaches are task-specific, and they rely heavily on prior knowledge and specially designed perception modules.

Due to the challenges in modeling bimanual coordination, recent work has increasingly shifted toward learning policies for bimanual manipulation from demonstration. Coupling low-cost hardware (e.g., ALOHA \cite{zhao2023learning} and Gello \cite{wu2024gello}) with generative imitation learning methods has made it more feasible to replicate fine-grained, coordination-intensive bimanual behaviors. To mitigate compounding error in long-horizon tasks, several approaches \cite{yu2024bikc, gao2024bikvil, ppi2025} introduce dual-arm keyposes as explicit or implicit goal conditions. Our skill-partitioning strategy is related to the heuristic keypose extraction used in these methods, but we encode task understanding through contact-based rules.
Skill decomposition also appears in augmentation-driven pipelines, such as DexMimicGen \cite{jiang2025dexmimicgen}, which extends MimicGen \cite{mandlekar2023mimicgen} to bimanual settings. In contrast to approaches that rely on large training corpora, we focus on the low-data regime, using as few as 20 demonstrations per skill to handle long-horizon, contact-rich bimanual manipulation tasks on real hardware.

\subsection{Techniques for Spatial Generalization} %
Methods that aim to generalize to random object placements fall into two main categories: 1) Generating more data that cover a larger range of space using data augmentation;
2) Exploiting the symmetry in data by specially designed equivariance models. We abbreviate them as augmentation methods and equivariance methods, respectively.  

Augmentation methods normally partition demonstrations into object-centric trajectories, and then apply transforms on both the trajectory and the object placement to create new demonstration instances. A representative approach is MimicGen \cite{mandlekar2023mimicgen}, which replays modified trajectories to expand datasets containing image observations. Because real-world rollouts require substantial manual effort, recent simulation-based augmentation methods~\cite{chen2025robotwin,yang2025novel} rely on photorealistic simulation and subsequently perform sim-to-real transfer. In contrast, 3D IL methods \cite{ze20243d,yang2024equibot,ke20243d} enable object-centric augmentation approaches \cite{xue2025demogen, zhou2025you} that  apply transformations to object point clouds and end-effector trajectories. While these augmentation strategies support spatial generalization for short-horizon skills, the discrepancy between synthesized and real-world data may accumulate in more complex tasks.

Equivariance methods form another line of research, aiming to inject symmetry into IL \cite{yang2024equibot,ryu2024diffusion,gao2024riemann} so that predicted actions transform consistently with object-level transformations. However, extending these approaches to contact-rich, fine-grained manipulation remains difficult because commodity sensors struggle to provide reliable depth measurements at close range. Although recent work \cite{hu20253d, wang2025practical} studies symmetry representations from RGB inputs, these methods are restricted to single-arm settings that rely on wrist-mounted cameras.
This highlights a broader challenge in applying equivariant models to general-purpose robot policies: local observations can provide an almost object-centric view that supports equivariance, but they sacrifice global context and are prone to occlusion during contact. 
{Given the symbolic structure of TAMP, \ours inherently provides an object-centric view of the target object. In addition, by separating the contact-rich phase from the object-centric primitives, \ours reduces the difficulty of equivariance learning, making it less data-demanding and more reliable.}

\begin{table*}[ht]
\centering
\caption{A collection of hierarchical planning-learning methods. The sensor modality in \textbf{bold font} indicates the requirement for real-time updates. }
\begin{tabular}{lcccc}
\toprule
\textbf{Method}  & \textbf{Observation Modality} & \textbf{Contact-involved Operations} & \textbf{Skill Switching Condition} & \textbf{Source of Plan Skeleton} \\
\midrule
PSL \cite{dalal2024plan} & \textbf{RGB}  + object pose & RL-learned policies & Predefined & Predefined \\
League \cite{cheng2023league}  & \textbf{Object pose} & RL-learned policies & Predefined & Predefined \\
STAP \cite{agia2023stap}   & \textbf{Object pose} & RL-learned policies & Policy intersection  & Predefined \\
LSP \cite{xue2024logic}  & \textbf{Object pose} & RL-learned policies & Policy intersection & TAMP \\
SPIRE \cite{zhou2024spire}  & \textbf{RGB} + object pose & IL\&RL-learned policies & Predefined & TAMP \\
NOD-TAMP \cite{cheng2023nod} & Depth & IL-learned contact poses & --- & Predefined \\
Logic-LfD \cite{zhang2024logic}  & \textbf{Object pose} & IL-learned controller & Predefined & TAMP \\
GFC \cite{mishra2024generative}  & \textbf{Object pose} & IL-learned policies &  Policy intersection &  Predefined  \\
HITL-TAMP \cite{mandlekar2023human}  & \textbf{RGB} + object pose & IL-learned policies & Predefined & TAMP \\
SkillMimicGen \cite{garrett2024skillmimicgen}  & \textbf{RGB} + object pose & IL-learned policies & Learned Predictor & Predefined \\
BLADE \cite{liu2024learning}   & \textbf{RGB} + depth & IL-learned policies & Learned Predictor & Task Planning \\
MT3 \cite{dreczkowski2025learning}   & RGB + depth & Retrieved trajectories & Predefined & Predefined \\
\rowcolor{gray!15}  \ours (Ours)  & \textbf{RGB} + depth & IL-learned primitives\&policies & Learned Predictor & TAMP \\
\bottomrule
\end{tabular}
\label{tab:learn-plan-literature}
\vspace{-0.15in}
\end{table*}

\subsection{Learning structural information from demonstrations and foundation models}

Several recent works convert task demonstrations, human intention, or foundation-model semantics into reusable structures for planning: contact-mode mechanisms from a single demo~\cite{mao2023learning}, predicates and operators from language feedback~\cite{han2024interpret}, planning domains from visual-language demonstrations~\cite{ye2025unidomain}, and demo-video-to-plan via VLM reasoning~\cite{wang2025vlm}. A parallel line uses foundation models for high-level skill sequencing~\cite{ahn2022can} or generates situated task plans via program-like prompting~\cite{singh2022progprompt}.

Our insight aligns with these works: foundation-model-level semantic understanding becomes reliably useful for robot deployment only after it is translated into structured, physically verifiable action representations. Like prior methods~\cite{mao2023learning,ye2025unidomain}, \ours converts demonstrations into plannable structure, but addresses the complementary robot-execution problem in three aspects. First, \ours turns demonstration data into learned, robot-executable skills that are exposed to TAMP as stream-backed components, including object-centric primitives and contact-rich visuomotor policies. Second, \ours verifies geometric feasibility through reachability and policy-safety constraints. These checks feed feasibility information back to the symbolic planner, which can then insert additional movements in the \texttt{unsafe} and \texttt{unreachable} scenarios described in Section~\ref{sec:constraint_handle}. Third, \ours closes the loop with online perception and replanning, which can be triggered when the contact detector identifies inconsistency between the observed state and the predicted state.

\subsection{Task and Motion Planning with Learned Operations} %

\subsubsection{Task and Motion Planning Methods}
Task and motion planning (TAMP) can compute solutions to complex manipulation problems that consist of various predefined motion primitives. This area is dominated by two popular paradigms: optimization-based TAMP and satisfaction-based TAMP.
Optimization-based TAMP methods \cite{toussaint2018differentiable,migimatsu2020object} refer to those optimizing both discrete and continuous variables with respect to a certain metric. 
On the other hand, satisfaction-based TAMP methods \cite{dantam2018incremental,garrett2018sampling, garrett2020pddlstream, pan2024task} formulate the problem as a constraint-satisfaction problem.  With a detailed environmental description and a comprehensive skill library, satisfaction-based TAMP can search for a set of discrete and continuous variables that comply with specified constraints and targets. Among them,  sampling-based TAMP methods \cite{garrett2018sampling} naturally combine purpose-agnostic sampling-based motion planners into their stream designs, making the whole system modular and extensible. 

However, the aforementioned TAMP methods often oversimplify contact poses, assuming predefined object-specific grasp poses are available.  
M0M \cite{curtis2022long} integrated open-set perception \cite{xie2021unseen} and general-purpose grasping modules \cite{gualtieri2016high,mousavian20196}  into TAMP, thereby relaxing the requirement of precise object geometry. However, pre-trained general-purpose grasping modules often fail to provide satisfactory contact poses in challenging cases, e.g., with thin objects or noisy depth measurements. Furthermore, even when the produced grasp poses are feasible, the resulting contact poses are usually distinct from those in demonstration data. %
In contrast, our approach directly mimics robot-object contacts from human demonstrations, resulting in observations that can successfully initiate an imitation policy trained from those demonstration data.

\subsubsection{Hierarchical Planning-learning Methods}
To compensate for weaknesses of traditional planning methods in robot-object interaction, there is increasing interest in combining learned operations with classical planning. Operations, either in the form of a contact pose or a policy, can be learned via reinforcement learning (RL) or imitation learning (IL). We list a series of related works in Table \ref{tab:learn-plan-literature}.

A significant body of work integrates RL-learned policies with symbolic plans. Methods like PSL \cite{dalal2024plan}, League \cite{cheng2023league}, STAP \cite{agia2023stap},  LSP \cite{xue2024logic}, and SPIRE \cite{zhou2024spire} typically use a high-level plan to sequence low-level skills, sometimes integrated with optimization-based TAMP \cite{xue2024logic} or sampling-based TAMP \cite{zhou2024spire}.  While effective, these approaches often depend on dense rewards from simulators \cite{cheng2023league} or require privileged information like object poses \cite{ agia2023stap, xue2024logic}. Although the visual policies of these methods~\cite{dalal2024plan, zhou2024spire} do not require ground-truth object poses in real time, they frequently face notorious sim-to-real transfer challenges in real-world deployment. That goes against our intention to devise a system that is easily deployed in the real world.

Another line of research relies on human demonstrations to bypass the need for dense rewards and sim-to-real transfer. Some methods integrate learned contact poses (NOD-TAMP \cite{cheng2023nod}) or reactive controllers (Logic-LfD \cite{zhang2024logic}) into TAMP, 
though contact-rich operations are excluded from their application. MT3 \cite{dreczkowski2025learning} adopts a decomposition idea similar to ours, separating a skill into alignment and interaction phases. The above methods are primarily restricted to single-interaction tasks, which differ from our applications with bimanual operations.
On the other hand,  HITL-TAMP \cite{mandlekar2023human} and its successor SkillMimicGen \cite{garrett2024skillmimicgen} integrate visuomotor policies with planned trajectories, producing more flexible behaviors. 
A common limitation for the majority of these IL-based approaches \cite{mandlekar2023human, garrett2024skillmimicgen, zhou2024spire, dreczkowski2025learning} is their reliance on 6-DoF object poses which are difficult to obtain reliably, especially for objects that exhibit symmetry \cite{liu2022gen6d, fan2024pope, wen2024foundationpose}. \ours avoids explicit 6-DoF object pose estimation by leveraging object-level equivariance.

A central challenge in such hybrid systems lies in defining the skill switching criterion: identifying the precise moment to halt one operation and initiate the next. Traditional methods often depend on rigid, handcrafted rules \cite{dalal2024plan, mandlekar2023human,dreczkowski2025learning}. Some approaches \cite{agia2023stap, xue2024logic, mishra2024generative} find a subgoal in the intersection of two policies by optimizing the value function or state distribution in low-dimensional space. However, it is rather difficult to estimate value functions and distributions with high-dimensional image observation. 
{BLADE \cite{liu2024learning} and SkillMimicGen \cite{garrett2024skillmimicgen} get rid of privileged information at test time by developing a visual task progress predictor. Unlike those~\cite{liu2024learning, garrett2024skillmimicgen} whose  predictors capture per-object state transitions from a global image view, our approach focuses on contact-centric cues using wrist-camera images,  which is easier to learn and robust to background noise.}

\section{Problem Formulation} %
Section \ref{sec:preliminary_tamp} reviews sampling-based TAMP, Section \ref{sec:preliminary_sg} extends the scene-graph structure for symbolic skill representation, and Section \ref{sec:preliminary_problem} defines the studied problems and gives a system overview.

\subsection{Sampling-based TAMP} \label{sec:preliminary_tamp}
We formulate the planning of hybrid learned-planned operations in PDDLStream \cite{garrett2020pddlstream} format, a popular sampling-based TAMP description format. PDDLStream is an extension to the Planning Domain Definition Language (PDDL) that supports planning with continuous values using conditional samplers. 
A PDDLStream problem  $\langle \textbf{X},  \mathbb{P}, s_0,\eta, \mathbf{A}, \Psi \rangle$ is given by a set of variables $\textbf{X}$,  a set of predicates $\mathbb{P}$, an initial state  $s_0  $, a goal description $\eta $, a set of skills $\mathbf{A}$, and a set of streams $\Psi$. 
A predicate $\varsigma \in \mathbb{P}$ is a Boolean function with respect to variables $ \textbf{X}$, representing a binary property of the system state. Variable set $\mathbf{X}$ includes both discrete entities (e.g., grippers, objects, skills, etc.) and continuous quantities  (e.g., trajectories, object poses, relative contact poses, etc.). 
The fact set $\Lambda$ is defined as the set of predicates instantiated from $\mathbb{P}$ with respect to symbolic variables $X$, and $\textbf{S} \triangleq 2^{\Lambda}$ is the symbolic state space.
A discrete skill $a \in \mathbf{A}$ has a precondition formula $\text{pre}(a) \subseteq  \Lambda $  and an effect formula $\text{eff}(a) \subseteq  \Lambda$. 
A skill $a$ is applicable in a state $s \subseteq \Lambda$ if and only if its preconditions are met, i.e.,   $\text{pre}(a) \subseteq s $. 
If a skill $a$ is applied in state $s$, it results in a new state $s' = (s \setminus \text{eff}^{+}(a)) \cup \text{eff}^{-}(a) $, where {$\{ \text{eff}^{+}(a), \text{eff}^{-}(a) \} \subseteq \Lambda $} are facts to add and delete.
Stream functions $ \Psi: \Lambda \rightarrow \Lambda$ are conditional generators that produce new facts conditioned on already grounded facts.  Typically, streams are implemented in two formats:  1) Sampling streams, which may output certified properties and valued continuous variables (e.g., computed trajectories); 2) Test streams, which only output a binary evaluation of a specific property (e.g., safety). 

\subsection{Contact-aware {Scene Graph}} 
\label{sec:preliminary_sg}

\begin{table}[t!]
\centering
\caption{Mapping rules in $\mathsf{EdgeOps2Predicates}(\cdot)$ function.}
\footnotesize
\label{tab:EdgeOps2Predicates}
\setlength{\tabcolsep}{3pt}
\begin{tabular}{@{}C{0.28\columnwidth}C{0.14\columnwidth}|L{0.48\columnwidth}@{}}
\toprule
\textbf{Edge}                       & \textbf{Operation} & \textbf{Predicates}                                 \\ \midrule
\multirow{2}{*}{$ \langle h,o \rangle$}      & Add        & $ \pddl{AtGrasp}(h,o,?g)$ \\ \cmidrule(lr){2-3}
                            & Del     & $ \neg \pddl{AtGrasp}(h,o,?g)$ \\ \midrule
\multirow{2}{*}{$ \langle o_2, o_1 \rangle, o_2 \in O_\mathrm{on}$} & Add        &   $\pddl{On} (o_1, o_2), \pddl{AtPose}(o_1, o_2,  ?p)$     \\ \cmidrule(lr){2-3}
                            & Del     &  $  \neg \pddl{On} (o_1, o_2),\neg \pddl{AtPose}(o_1, o_2, ?p)$  \\  \midrule
    $\langle o_2, o_1 \rangle, o_2 \in O_\mathrm{in}$ & Add & $\pddl{In} (o_1, o_2)$  \\
\bottomrule
\end{tabular}
\vspace{-0.1in}
\end{table}

\begin{figure}[t!]
    \centering
    \includegraphics[width=0.9\linewidth]{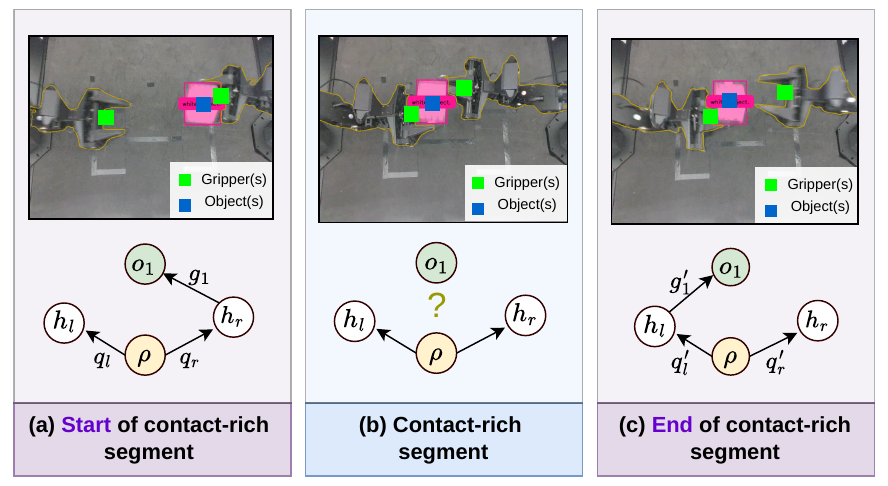}
    \caption{(a) and (c) show the \textit{start} and \textit{end} of the contact-rich segment of the \textit{Object Handoff} operation; (b) shows an intermediate contact-rich state. Scene graphs are shown below each frame. States matching the precondition graph $G_\mathrm{pre}$ in (a) and the effect graph $G_\mathrm{eff}$ in (c) are used for skill transitioning, while intermediate states are used for skill replication. Symbols are world frame $\rho$, object $o_1$, left arm $h_l$, right arm $h_r$, initial arm configurations $(q_l, q_r)$, initial contact pose $g_1$, final arm configurations $(q'_l, q'_r)$, and final contact pose $g'_1$. 
    }
    \label{fig:transfer_sg}
    \vspace{-0.18in}
\end{figure}
We model contact relations as graph edges and use a scene-graph representation for the symbolic state space. In our tabletop manipulation setting, a scene graph is a directed graph ${G} = {V, E}$. The node set $V = {\mathbf{H}, \mathbf{O}, \rho}$ includes robot grippers $\mathbf{H}$, objects $\mathbf{O}$, and the world frame $\rho$. Each object belongs to one or more of the following subsets: $O_\mathrm{on}$ (objects with supporting surfaces), $O_\mathrm{in}$ (containers), $O_\mathrm{movable}$ (prehensile manipulable objects), and $O_\mathrm{fixed}$ (immobile objects).
The edge set $E \subseteq V \times V$ encodes fixed transforms or kinematic links, where the first node in an edge is the parent. For instance, an edge $\langle o_2, o_1 \rangle$ indicates that object $o_2$ \textit{supports} object $o_1$, and an edge $\langle h, o \rangle$ indicates that gripper $h$ \textit{holds} object $o$. Finally, we define $\tilde{E} = {(e, op) \mid e \in E,; op \in {\mathrm{Add}, \mathrm{Del}}}$ as the set of edge operations corresponding to establishing or breaking contacts.
We define a function, $\mathsf{EdgeOps2Predicates}(\cdot)$, to map an edge operation $\tilde{e} \in \tilde{E}$ to predicates. 
The detailed mapping can be viewed in Table \ref{tab:EdgeOps2Predicates}, where $\neg$ denotes negation. Notably, the mapping is also conditioned on object properties. For example, if one object $o_1 \in O_\mathrm{movable}$ is dropped inside a container-like object $o_2 \in O_\mathrm{in}$,  $o_1$ is considered \textit{absorbed} by $o_2$ \cite{agostini2023unified} and no positional predicates about $o_1$ are created.
All free variables are marked with `$?$', which  will be grounded during TAMP computation. 

Fig. \ref{fig:transfer_sg} showcases an example of graph abstraction to a bimanual handing-off skill.
We denote the graph abstraction of the starting state of a skill as a precondition graph $G_\mathrm{pre}$, and likewise, the graph abstraction of an end state is referred to as an effect graph $G_\mathrm{eff}$. We name the graph representations $G_\mathrm{pre}$ and $G_\mathrm{eff}$ as the \textit{contact signature} of a skill that abstracts away the possible stochastic behavior in intermediate states.
We assume that for a common gripper-object edge $\langle h, o \rangle$ across different skills, the relative contact poses of that edge always lie in a similar distribution.
We assume that the contact pose distribution is unimodal, so that  different learned skills from different demonstration sets can be directly sequenced according to their contact signatures. If we consider more than one contact mode for the same edge $\langle h, o \rangle$, we should introduce an additional contact mode argument $m$ in \texttt{Grasp}($h,o,m,g$). Otherwise, an incompatible contact mode will result in OOD wrist-camera observations for the subsequent learned policy, causing the task to fail.

\begin{figure*}[ht!]
    \centering
    \includegraphics[width=0.8\linewidth]{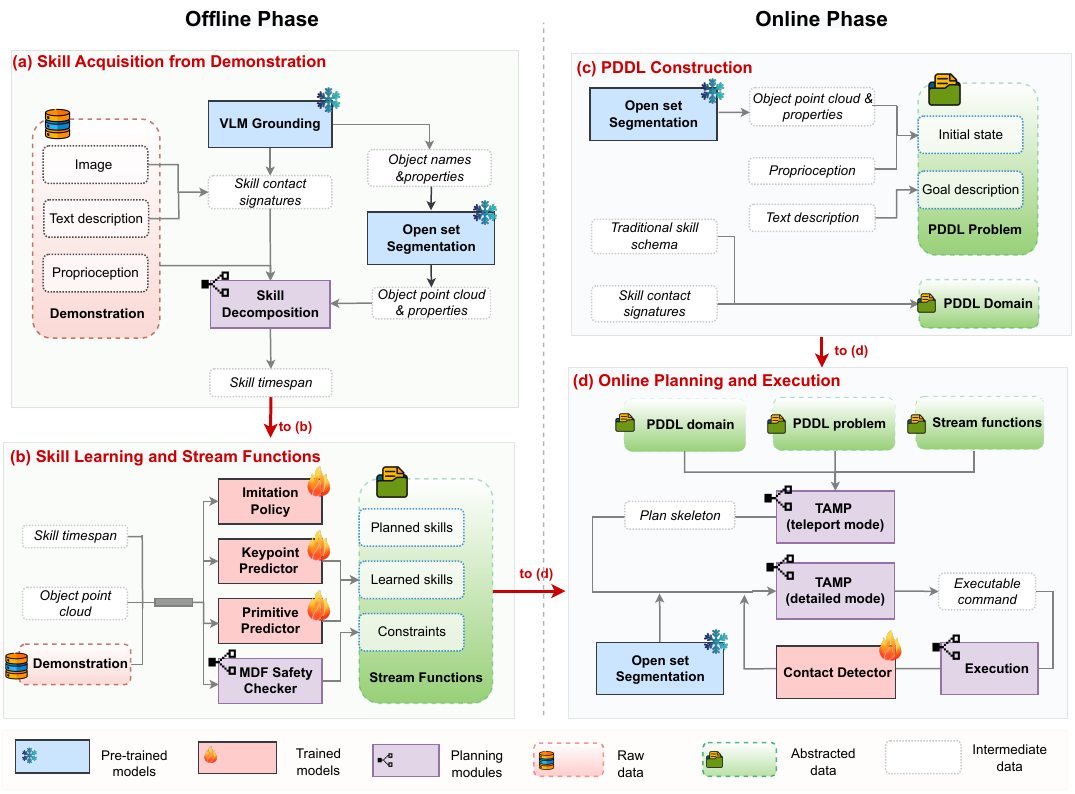}
    \caption{The overall diagram of \ours, organized into offline and online phases. In the offline phase, (a) skill acquisition and decomposition extract task objects, contact signatures, and skill timespans from demonstrations, and (b) the corresponding segments are used to train primitives, policies, keypoint predictors, and safety checkers that are wrapped as stream functions. In the online phase, (c) current observations and skill signatures are converted into the PDDL problem and domain, and (d) TAMP plans, executes, and verifies skills through contact-triggered perception. }
    \label{fig:overall-diagram}
    \vspace{-0.2in}
\end{figure*}

\subsection{TAMP with Skills Learned from Demonstrations} \label{sec:preliminary_problem}

We aim to solve a \textit{Decompose-and-Reorganize} problem for general long-horizon and real-world manipulation tasks: 1) split demonstrations into atomic skills, and 2) reassemble skills into a plan that can accomplish potentially out-of-distribution tasks.
The skill-acquisition problem and the plan composition problem can be defined as Problem \ref{prob:skill-acquiring} and Problem \ref{prob:plan-composition}, respectively.
\begin{Problem}
\label{prob:skill-acquiring}
Given the task description and a set of demonstrations, 
construct a repertoire of contact-involved skills for sampling-based TAMP.  
\end{Problem}
\begin{Problem}
\label{prob:plan-composition}
Given a skill repertoire, compute a plan sequence that can replicate a demonstrated task or a new task composed of skills in the repertoire. 
\end{Problem}

To address these two core problems, we propose \ours, whose overall workflow is summarized in Fig.~\ref{fig:overall-diagram}. Panels (a) and (b) show the offline construction of the skill repertoire: demonstrations are decomposed into contact-aware skill timespans (Section \ref{sec:skillseg}), and each skill is learned or wrapped into stream functions (Section~\ref{sec:streams}). Panels (c) and (d) show the online reorganization stage: current observations instantiate the PDDL problem and domain (Section~\ref{sec:schemata}), and the planner connects stream outputs to execution, contact verification, perception, and local replanning (Section \ref{sec:online}).

\section{Skill Acquisition} \label{sec method} %

This section presents the decomposition stage: a VLM-assisted, contact-aware approach segments and categorizes demonstrations into skills (see {Section~\ref{sec:skillseg}}).
According to skill semantics and contact complexities, we divide skills into three distinct categories: learned object-centric primitives, learned visuomotor policies, and planned trajectories.
Specifically, for segments with lower behavior complexity (e.g., pick, place, drop), we use an equivariant network to learn open-loop object-centric primitives  (\textit{see Section~\ref{sec:primitivelearn}}). 
For challenging contact-rich segments (e.g., wiping, flipping, folding), we replicate them by learning a visuomotor policy, and predict keyposes at both ends of each skill for planning (see {Section~\ref{sec:policy}}).
When no contact is involved, motion planning takes charge of the segment, ensuring seamless integration of different skills. 
Finally, {Section \ref{sec:streams}} introduces the procedure of encoding learned skills into a uniform symbolic representation, i.e., stream functions, which can be directly plugged into modern sampling-based TAMP systems. 

\begin{figure}[ht]
    \centering
    \includegraphics[width=\linewidth]{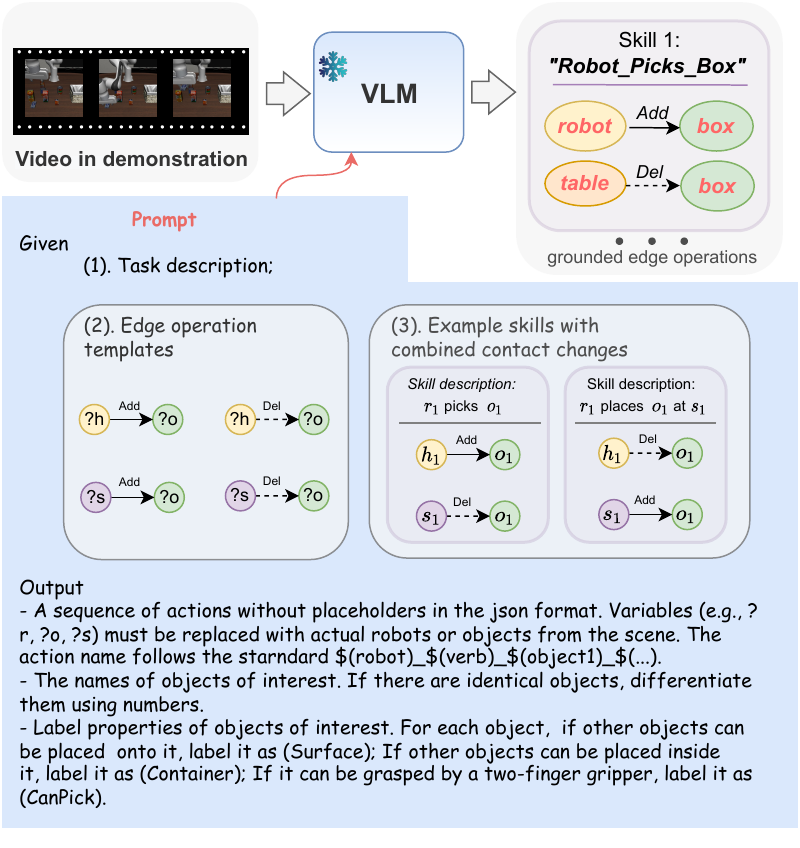}
    \vspace{-0.25in}
    \caption{Skill detection and scene grounding.}
    \label{fig:vlm-grounding}
    \vspace{-0.2in}
\end{figure}

\subsection{Skill Detection, Abstraction, and Partition} \label{sec:skillseg} %

We develop a VLM-assisted and contact-aware approach to perform task understanding: a VLM grounding module first abstracts graph-like representations of atomic skills from demonstrations, and a contact-aware module conducts fine-grained skill segmentation with edge-operation guidance.

\subsubsection{VLM-assisted Grounding}
The workflow and prompt design are shown in Fig.~\ref{fig:vlm-grounding}. 
By design, we provide the \textit{VLM (i.e., Qwen-VL-Max)} with a limited set of edge operations, representing gripper-object or object-object interactions. 
Additionally, to better integrate and associate contact changes with skills, we supplement the prompt with skill examples, e.g., pick and place, as in-context learning material. 
The VLM is expected to use video information to ground the static variables in edge operations and detect a sequence of discrete skills $\{a_i\}_{i=1}^n$, where each skill is composed of grounded edge operations $\{\tilde{E}_i\}_{i=1}^n$. 
Rather than asking the VLM to output skills and their contact edges in a single pass, we prompt it for two coupled outputs: an action-label sequence in execution order, and a set of contact edge operations in which each operation is tagged with the skill it belongs to.
Forcing every edge operation to reference a previously listed skill makes the VLM commit to the skill semantics before detecting any contact, which reduces hallucination. A rule-based step then assembles the tagged operations into each skill's contact changes.
Furthermore, the VLM  also captures the names and properties of task-related objects from the input video and task description, and labels each task-related object with unary predicates. For reliability purposes, we manually check the correctness of VLM output before the next data processing step.
As all demonstration episodes share the same scene and skill sequence, the above VLM-assisted grounding procedure is performed once per task. The statistical analysis of VLM-assisted grounding is further analyzed in Appendix~\ref{sec:appendix-grounding-failures}.

\begin{algorithm}[t!] \footnotesize
\caption{$\mathsf{GraphAbstraction}(\cdot)$}\label{alg BuildSubGraph}
\LinesNumbered %
\KwIn{Arm $\mathbf{H}$, the world frame $\rho$,  skill sequence $\{a_i \}^n_{i=1}$, edge operations $\{\tilde{E}_i \}^n_{i=1}$ }%
\KwOut{Updated skill sequence $\{a_i \}^n_{i=1}$}%
$V, E_0 \gets \mathsf{InitialGraph} (\{\tilde{E}_i\}^n_{i=1})$ \\ \label{line:initialgraph}
\For{$i \in 1,\dots, n $}{
\brown{\tcp{label skill with precondition graph and effect graph}}
$H_i \gets \mathsf{InvolvedArms}(a_i, \mathbf{H})$ \\ \label{line:InvolvedArms}
$a_i.G_\mathrm{pre} \gets  \mathsf{Subgraph}(V, E_{i-1}, H_i, \rho)$ \\ \label{line:SubGraph1}
$E_{i} \gets \mathsf{ApplyEdgeChanges} (E_{i-1}, \tilde{E}_i)$ \\ \label{line:applyedgechanges}
$a_i.G_\mathrm{eff} \gets \mathsf{Subgraph}(V, E_{i},H_i, \rho)$ \\ \label{line:SubGraph2}
}
\Return $\{a_i \}^n_{i=1}$
\end{algorithm}

\subsubsection{Graph Representation for Skills}
Each VLM-detected skill is described in the \texttt{$\$(\text{robot})\_\$(\text{verb})\_\$(\text{object1})$} format, for instance, \texttt{robot\_picks\_box} (see Fig. \ref{fig:vlm-grounding}).  We denote the text description of a skill $a$ as $\lambda_a$ in the rest of this paper. 
In addition,  for each skill, Algorithm~\ref{alg BuildSubGraph} introduces the process to link  the starting and ending contact states with their corresponding graph interpretation, denoted as $G_\mathrm{pre}$ (precondition graph) and $G_\mathrm{eff}$ (effect graph), respectively.
Specifically, in \textit{Line \ref{line:initialgraph}}, the function $\mathsf{InitialGraph}(\cdot)$ infers the set of edges $E_0$ present in the initial scene graph  $G_0 = (V, E_0)$ by analyzing the grounded edge operations $\{ \tilde{E}_i \}^n_{i=1}$ output by VLM. The principle of defining $E_0$ is based on the first observed operation for any given edge $e$ in the sequence: 
If the first operation applied to an edge $e$ is $\mathrm{Add}$, then $e$ must have been absent from $G_0$; 
If the first operation applied to an edge $e$ is $\mathrm{Del}$, then $e$ must have been present in $G_0$.
After the establishment of $G_0$, a loop iterates over each skill $a_i$ and sequentially applies edge changes $\tilde{E}_i$ to $G_0$ with $\mathsf{ApplyEdgeChanges}(\cdot)$. In each iteration,  the current edge set $E_{i-1}$ is used to derive the precondition graph $a_i.G_\mathrm{pre}$ (\textit{Line} \ref{line:SubGraph1}), and the next edge set $E_{i}$ is used to derive the effect graph $a_i.G_\mathrm{eff}$ (\textit{Line} \ref{line:SubGraph2}). 
{Notably, we use $\mathsf{InvolvedArms}(\cdot)$ (see \textit{Line} \ref{line:InvolvedArms}) and  $\mathsf{SubGraph}(\cdot)$  (see \textit{Line \ref{line:SubGraph1} / \ref{line:SubGraph2}}) to cope with parallel execution of unimanual skills in a dual-arm manipulator. In detail, let the graph $\bar{G}$ be the graph that includes all entities except the static world frame $\rho$. Each arm belongs to a connected component in $\bar{G}$, which is identified as a subgraph using $\mathsf{Subgraph}(\cdot)$. Thus, when the skill $a_i$ involves only one arm, its graph interpretations  $a_i.G_\mathrm{pre}$ and  $a_i.G_\mathrm{eff}$ remain unaffected by the task progress of the other arm.}
For skills containing multiple contact changes, we treat the edge operations as an unordered set associated with the same skill.
This abstraction preserves flexibility during timespan partition, allowing variation of the contact order across demonstrations.
For example, in a dropping skill, object-object contact may be established before or after the gripper-object contact is released, yet both cases induce the same effect graph.

\subsubsection{Contact-aware Timespan Partition}
We partition the timespan of demonstration episodes into distinct phases for skill training.
Specifically, a skill switching point lies between two contact change events, which are detected by object-object or gripper-object distances: the contact is established if the closest distance is below a predefined threshold, and broken otherwise.
The object point cloud is directly available in simulated tasks, while in real-world tasks,  \textit{GroundedSAM  \cite{ren2024grounded}}  and \textit{XMem} \cite{cheng2022xmem} obtain object masks and track them across the video,  and the point cloud is extracted at each timestep.
For the gripper point clouds, we employ a \textit{point cloud forward kinematics model \cite{wei2024d}} to obtain them at each timestep. 
Finally, each graph-structure contact signature is associated with its timespan to form a distinct, trainable skill segment (e.g, the `initiation', `intermediate', and `termination' phases in Fig. \ref{fig:transfer_sg}).

\begin{figure}[t]
\centering
\includegraphics[width=\columnwidth]{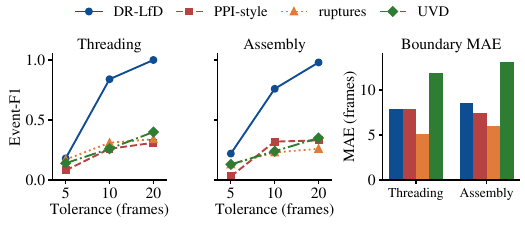}
\caption{Event-level comparison on two DexMimicGen tasks. Each prediction is matched to the closest annotated event, while unmatched predictions count as false positives. Event-F1 measures whether a predicted  event falls within tolerance windows of 5, 10, and 20 frames of the ground truth event. Boundary MAE is the mean absolute temporal error over matched events.}
\label{fig:event_seg}
\end{figure}

For trajectory planning purposes, \ours decides the attaching/detaching moment of each primitive by computing an \textit{in-hand status}, which is decided by whether the velocity cosine similarity  of an object aligns with that of the gripper. 
This is a part of our contact-aware partitioning strategy, which is simple but achieves higher Event-F1 than alternatives like PPI-style keypoints~\cite{ppi2025}, UVD~\cite{zhang2024universal}, and \texttt{ruptures}~\cite{truong2020selective}. Fig.~\ref{fig:event_seg} reports Event-F1 and boundary MAE on the DexMimicGen \textit{Two-arm-threading} and \textit{Two-arm-assembly} tasks: \ours achieves higher Event-F1 by selecting one event per primitive to reduce false positives.

In summary, the proposed VLM-assisted and contact-aware segmentation approach is tightly linked to the geometric part of the symbolic  TAMP representation.
For example, the graph-structure contact signatures and edge operations are interpreted as predicates, serving as symbolic skill representation (see Section~\ref{sec:schemata}) and subgoals (see Section \ref{sec:online}).

\subsection{Learning Object-centric Primitives } \label{sec:primitivelearn} %

\begin{figure}[t!]
    \centering
    \includegraphics[width=1\linewidth]{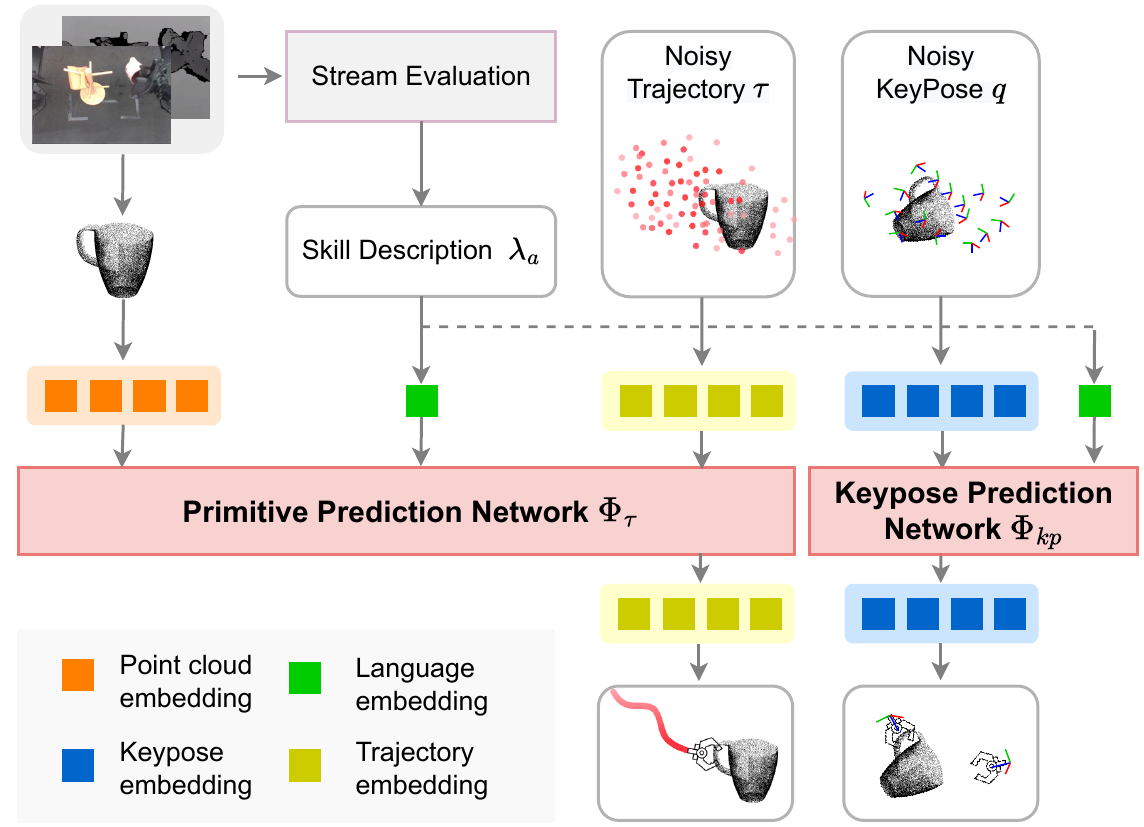}
    \caption{The primitive prediction network $\Phi_\tau$ and the keypose prediction network $\Phi_{\mathrm{kp}}$. }
    \label{fig:so3-skill}
    \vspace{-0.2in}
\end{figure}

After demonstration decomposition performed in Section \ref{sec:skillseg}, we train networks to reproduce the trajectory or keyposes of each skill, as shown in Fig.~\ref{fig:so3-skill}.
Following the skill categorization in Section~\ref{sec method}, we train an \textit{object-centric primitive predictor} $\Phi_{\tau}$ for low-complexity skills, and, for contact-rich skills, a \textit{visuomotor policy} $\pi$ together with a \textit{keypose predictor} $\Phi_\mathrm{kp}$. The predictor $\Phi_{\tau}$ outputs a complete motion trajectory, whereas $\Phi_\mathrm{kp}$ outputs only the initial and target keyposes without intermediate trajectories. 
This section focuses on the methodology of primitive learning, while Section \ref{sec:policy} introduces the details of training the policy and keypose predictors. 

The network $\Phi_{\tau}$ is designed for learning a skill in the form of an open-loop object-centric primitive $\tau$, which comprises three trajectory types: a trajectory of end-effector pose in $SE$(3), a trajectory of the $d$-joint fingers in $\mathbb{R}^d$, and a trajectory of in-hand status introduced in Section~\ref{sec:skillseg}.

Fig. \ref{fig:detail-so3} makes a deep dive into the network structure of $\Phi_\tau$, which consists of two main components: \textit{observation encoding network}  and \textit{noise prediction network}. 
To learn skills that generalize to novel placements, we employ \textit{Vector Neurons} \cite{deng2021vector} to achieve \textit{SO}(3)-equivariance in both the encoding network and noise prediction network. 

\subsubsection{Encoding of Multi-modal Observations} %
\begin{figure*}[tb]
    \centering
    \includegraphics[width=0.83\linewidth]{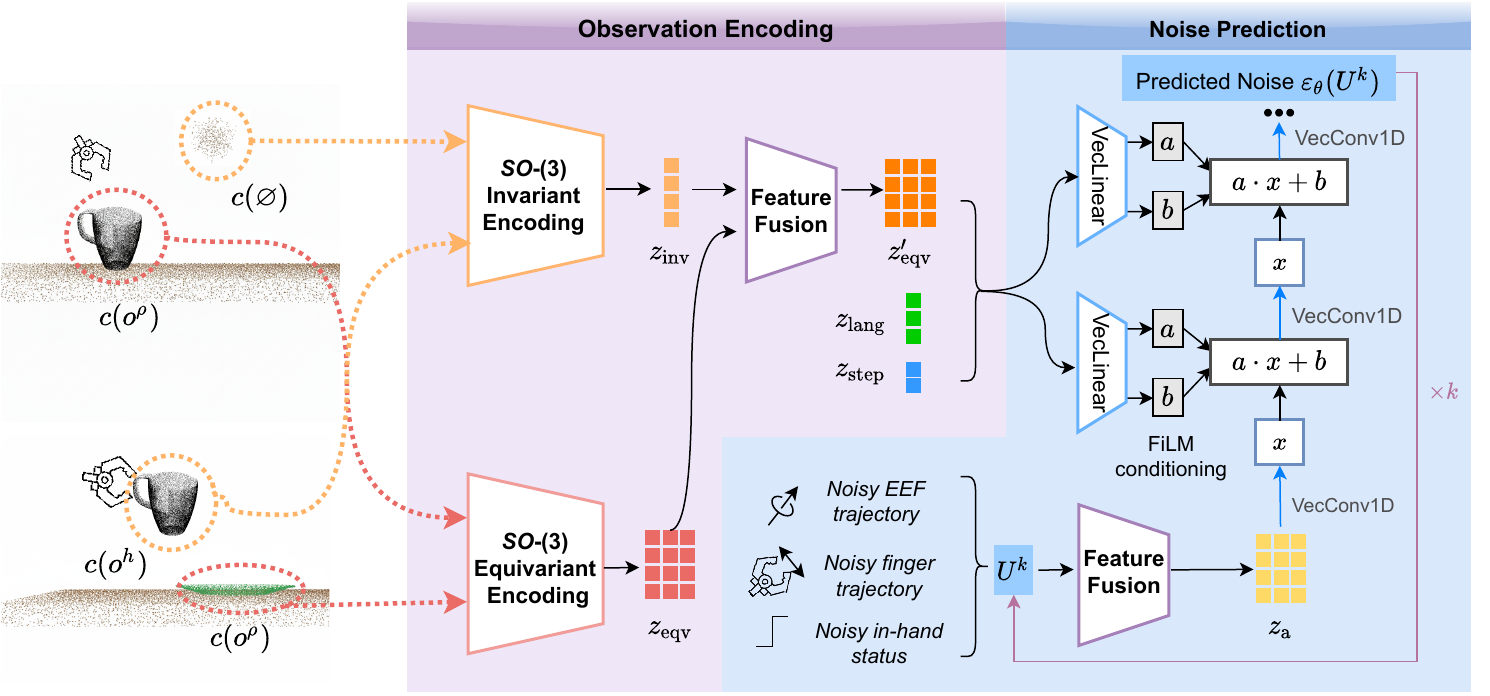}
    \caption{The \textit{SO}(3) equivariant network for primitive learning, containing the observation encoding part and noise prediction part. The output $U$ contains an object-centric end-effector (EEF) trajectory, the finger trajectory in a scalar array, and the index when a contact is established between a pair of entities. }
    \label{fig:detail-so3}
    \vspace{-0.18in}
\end{figure*}

The noise prediction network takes the first-sight point cloud observation $ \{C_{a_i}\}_{i=1}^M$ and skill-related language descriptions $\{\lambda_{a_i} \}_{i=1}^M $ as conditions, where $M$ is the number of primitives. The language input is tokenized into embeddings $z_{\text{lang}}$ by BERT \cite{devlin2019bert}. On the other hand, the point cloud observed at the first frame of each primitive is processed by the \textit{VN-DGCNN encoder} \cite{lei2023efem} in  Figure \ref{fig:detail-so3}.

For illustration, consider the contact signature $G_\mathrm{pre}$ of the placement skill shown in the lower-left of Fig.~\ref{fig:detail-so3}.  
Let the held object (e.g., a cup) be $o^h$, the object it is placed onto (e.g., a plate) be $o^\rho$, the point cloud of object $o$ be $c(o)$, and the \textit{VN-DGCNN} encoder be $\mathsf{Enc}(\cdot)$.
The encoder maps the held object to a rotation-invariant feature $z_\mathrm{inv} = \mathsf{Enc}(c(o^h))$, which captures geometric information that does not vary under rotations.  
In parallel, it maps the static object $o^\rho$ to a rotation-equivariant feature $z_\mathrm{eqv} = \mathsf{Enc}(c(o^\rho))$, satisfying the equivariance property $\mathsf{Enc}(R \cdot c(o^\rho)) = R \cdot z_\mathrm{eqv}$ for any rotation $R$.

We employ an \textit{object-level padding technique} to accommodate skills with different graph representations $G_\mathrm{pre}$, such as the pick skill in the top-left of Fig. \ref{fig:detail-so3}. Specifically,  for the empty arm before picking, a dummy held object point cloud $c(\varnothing)$ is created. 
The \textit{SO}(3)-equivariant feature fusion module  $f_\mathrm{fuse}$ outputs another \textit{SO}(3)-equivariant vector $z'_\mathrm{eqv} = f_\mathrm{fuse}(z_\mathrm{eqv}, z_\mathrm{inv})$. By construction, $f_\mathrm{fuse}$ is equivariant under any rotation applied to $o^\rho$, i.e., $f_\mathrm{fuse}(\mathsf{Enc}(R \cdot c(o^\rho)), z_\mathrm{inv}) = R \cdot z'_\mathrm{eqv}$. After applying $f_\mathrm{fuse}$ to all held objects, the final point cloud embedding $z'_\mathrm{eqv}$  contains geometric information of all objects in $G_\mathrm{pre}$. With the padding technique, all skills can be trained using a single network.

\subsubsection{Learning Trajectory Distribution} %

We model each atomic skill $a_i$ as conditionally independent given its point cloud observation $C_{a_i}$ and language description $\lambda_{a_i}$. Under this assumption, the distribution of the trajectories of $M$ primitives can be factorized as:
\begin{equation}
\label{eq:maxprob}
\begin{aligned}
    P( \{\tau_{a_i}\}_{i=1}^M
| \{C_{a_i}\}_{i=1}^M, \{\lambda_{a_i} \} )= 
    \prod^{M}_{i=1} P(\tau_{a_i}|C_{a_i}, \lambda_{a_i}).
\end{aligned}
\end{equation}
We then leverage the data fitting power of Denoising Diffusion Probabilistic Models (DDPM) \cite{ho2020denoising} to learn the distribution of trajectories conditioned on the point cloud and language description. 
We train a single, unified score function, $\varepsilon_\theta$, on a dataset of individual skill tuples $\{ (\tau_{a_i}, C_{a_i}, \lambda_{a_i}) \}$, which learns to approximate the score function of this marginal distribution $P(\tau_{a_i}|C_{a_i}, \lambda_{a_i})$.
The noise predictor $\varepsilon_\theta(x,k)$ approximates a negative multiple of the score $\nabla_x \log P_k(x)$~\cite{salimans2021should,song2019generative,song2020scorebased}, where $k$ is the diffusion step and $P_k$ is the probability density of the noisy sample at step $k$.
Thus, the joint score function $\hat{\varepsilon}$ is the concatenation  of the individual score functions:
\begin{equation}
\label{eq:combinescore_matrix}
\begin{aligned}
\hat{\varepsilon}(\{\tau_{a_i}\}_{i=1}^M
| \{C_{a_i}\}_{i=1}^M, \{\lambda_{a_i}\}_{i=1}^M )
=
\begin{bmatrix}
\varepsilon_{\theta}(\tau_{a_1},k | C_{a_1},\lambda_{a_1})\\
\vdots\\
\varepsilon_{\theta}(\tau_{a_M},k  | C_{a_M},\lambda_{a_M})
\end{bmatrix},
\end{aligned}
\end{equation}
where $\varepsilon_\theta$ is an approximate score function for a single primitive.
The design of the noise prediction network  $\varepsilon_\theta$ follows that in Equibot \cite{yang2024equibot}, where a VNN-version U-Net is implemented. 
The optimal noise prediction network is trained by minimizing the diffusion loss \cite{ho2020denoising}.

Besides learning the \textit{SE}(3) end-effector pose trajectory, the network learns two additional scalar trajectories: 1) the gripper actions; 2) the contact states. The gripper actions are copied from the robot proprioception, while the binary contact indicator is labeled on each timestamp according to the contact changes detected in  Section \ref{sec:skillseg}. The timestamp of contact change $t_{cc}$ is used to compute approximate grasp poses $g$ and placed poses $p$. The scalar values are combined with vector types in \textit{SE}(3) by another feature fusion module, as pictured in Fig.~\ref{fig:detail-so3}.

Notably, in addition to the rotation-equivariance property of the backbone networks (i.e., both observation encoder and noise prediction network), the canonicalization of per-object coordinate enables translation-equivariance. This ensures that the learned trajectories are equivariant to arbitrary homogeneous transformations in the \textit{SE}(3) space.
\subsection{Learning Visuomotor Policies and Keyposes} \label{sec:policy} %

A skill involving contact-rich segments, e.g., bimanual operations, is learned using two networks including a \textit{visuomotor policy} $\pi$ and a \textit{keypose predictor} $\Phi_\mathrm{kp}$. 
Owing to the modular nature of our design, users can select arbitrary IL algorithms to train policies. 
The imitation policy is a learned mapping from system observation $\mathcal{O}$ to the low-level control signal $u = \pi (\mathcal{O}) $, where the system observation $\mathcal{O}$ typically contains robot proprioception $q$, wrist camera observation $I_\mathrm{wrist}$, and $3$rd-view camera observation $I_{3rd}$. 
One caveat is that current visual IL methods \cite{zhao2023learning,chi2023diffusion} are only applicable when the deployment observation has been \textit{encountered} in their training data. This implies that, in our hybrid planning-learning system, the end of a trajectory should produce an in-distribution observation so that a policy can follow. To this end, we design a \textit{keypose predictor} $\Phi_\mathrm{kp}$ to produce admissible initiation and termination conditions for each segmented skill.

Specifically, for a particular contact-rich skill $a$ and its contact signature $G$, keyposes are defined as proprioception data $q_{a.G}$ during the timespan of a contact signature $a.G$.
We learn the distribution of  keyposes in both initial and terminal contact states, i.e.,  $P(q_{a.G_\mathrm{pre}}, q_{a.G_\mathrm{eff}} \mid  \lambda_{a} )$,  conditioned on the skill description $\lambda_{a}$.   
Due to the stochastic nature of imitation policies, we opt to model the distribution of keyposes using a diffusion model, whose denoising network is a light-weight MLP. 
During inference, given the text description $\lambda_{a}$ provided by TAMP, the keypose predictor $\Phi_{\mathrm{kp}}$ yields the corresponding initial and terminal keyposes.
The predicted keyposes will be assembled by TAMP, marking the initiation and termination of a \textit{visuomotor policy} $\pi$.

\subsection{Learned Skills as Streams}  \label{sec:streams} %

This section introduces our design to encode the learned primitives and keyposes into stream functions, a form that is compatible with the PDDLStream \cite{garrett2020pddlstream} formulation. 
Traditional stream functions wrap sampling-based planners into conditional samplers,  certifying evaluated facts and generating valued variables, such as poses and trajectories.  
Appendix \ref{sec:planned_actions} summarizes the planned skills,  such as \textit{transit} (moving in free-space) and \textit{transfer} (moving while grasping an object), whose stream implementation is adopted from classical TAMP systems \cite{garrett2018sampling, vu2024coast}.
On the other hand, for VLM-detected skills, we design several network-integrated stream functions, which are mapped to those learned skills in Section \ref{sec:schemata} according to contact patterns. 

We implement the network-integrated stream functions shown in Table~\ref{tab:stream_tab}, following the notation from Fig.~\ref{fig:transfer_sg} (where `$?$' marks free variables, $\tau$ is a trajectory, and $\kappa$ is a policy identifier). Our key addition, departing from traditional PDDLStream formats, is the \textbf{Contact Pattern} column, which provides the explicit logic for the $\mathsf{MatchStreams}(\cdot)$ process (Section~\ref{sec:schemata}). This matching process follows a specific hierarchy: first, it determines if a skill is a visuomotor policy. If so, it is assigned $\mathsf{LearnedUniKeyPose}(\cdot)$ or $\mathsf{LearnedBiKeyPose}(\cdot)$ based on the number of arms. 
Then, if this policy's effect graph $G_\mathrm{eff}$ contains a gripper-object edge, it is also linked to the $\mathsf{LearnedPostGrasp}(\cdot)$ stream. On the other hand, if the skill is an object-centric primitive, it is instead matched to either $\mathsf{LearnedAttach}(\cdot)$ or $\mathsf{LearnedDetach}(\cdot)$ based on its \texttt{Add} or \texttt{Del} contact pattern. Note that due to space limitations, the table omits the full list of certified facts. Below, we break down the implementation details for each type.

\subsubsection{Primitive-relevant Streams}
{

The streams $\mathsf{LearnedAttach}(\cdot)$ and $\mathsf{LearnedDetach}(\cdot)$ correspond to the learned object-centric primitives, each involving a single contact change.
They rely on the same primitive prediction network $\Phi_\tau$ to compute a trajectory $\tau$, where each element comprises an end-effector (EEF) pose, a finger joint angle, and an in-hand status. The difference between $\mathsf{LearnedAttach}(\cdot)$ and $\mathsf{LearnedDetach}(\cdot)$ lies in their output variables. Specifically, $\mathsf{LearnedAttach}(\cdot)$ computes a contact pose $g$ defined as the pose of object $o$ relative to the EEF frame, while $\mathsf{LearnedDetach}(\cdot)$ computes the world-frame pose $p$ once the placed object $o_1$ is detached from the EEF.
Given an object pose $p^o$, a predicted primitive $\tau$ containing EEF trajectory $(p^{eef}_{0}, \dots, p^{eef}_n)$ and an index $i$ of the rising edge of the predicted in-hand status, $\mathsf{LearnedAttach}(\cdot)$ calculates the contact pose as $g = (p^{eef}_i)^{-1} \cdot p^o$.
The calculation of the placed pose $p$ in $\mathsf{LearnedDetach}(\cdot)$ follows a similar procedure.
Notably, the computed contact pose $g$ and placed pose $p$ are not required to be perfectly accurate in the global TAMP coordinate system, as the online perception (Section~\ref{sec:online}) will correct this geometric information after each learned skill is executed.}

\subsubsection{Policy-relevant Streams}
The stream functions $\mathsf{LearnedUniKeyPose}(\cdot)$ and $\mathsf{LearnedBiKeyPose}(\cdot)$ are designed to output the keyposes (e.g., $q, q'$) predicted by $\Phi_{\mathrm{kp}}$,  which serve to initiate or terminate the unimanual or bimanual visuomotor policies. 
{The symbol $\kappa$ denotes a \textit{policy identifier}, a data structure that encapsulates both execution logic and safety checks for a visuomotor policy. During planning, the test functions associated with each policy identifier $\kappa$ evaluate the potential collisions, as described in Section~\ref{sec:constraint_handle}. During execution, a policy identifier $\kappa$ triggers the visuomotor policy rollout when the state satisfies the conditions of $G_\mathrm{pre}$, and terminates the policy once the subgoals corresponding to $G_\mathrm{eff}$ are met. Details of the policy–primitive switching mechanism are provided in Section~\ref{sec:online}. 
}

The primitive-relevant and keypose streams together connect object-centric primitives with visuomotor policies. The primitive streams move objects into in-distribution  contact states, and the keypose streams then bring the robot proprioception to admissible policy initiation states. With both in place, TAMP can sequence a primitive before a policy, so the policy itself need not handle arbitrary out-of-distribution object placements. Appendix~\ref{sec:appendix-spatial-boundary} further discusses this mechanism and its capability boundary.

\begin{table}
    \centering
    \caption{Network-integrated stream functions and their input variables, output variables, and contact patterns.}
    \begin{tabular}{@{}c@{\hspace{6pt}}c@{\hspace{6pt}}c@{\hspace{6pt}}c@{}}
        \toprule
        \textbf{Name}          & \textbf{Inputs}                          & \textbf{Outputs}                      & \textbf{Contact Pattern} \\
        \midrule
        $\mathsf{LearnedAttach}(\cdot)$   & $(?h, ?o, ?p, ?a)$                       & $(?g, ?q, ?q', ?\tau)$                  & $(\langle ?h,?o \rangle, \text{Add})$ \\
        $\mathsf{LearnedDetach}(\cdot)$   & \makecell{$(?h, ?o_1, ?o_2,$ \\ $?a, ?g)$} & $(?p, ?q, ?q', ?\tau)$                  & \makecell{$(\langle ?h,?o_1 \rangle, \text{Del})$, \\ $(\langle ?o_2,?o_1 \rangle, \text{Add})$} \\
        $\mathsf{LearnedUniKeyPose}(\cdot)$    & $(?h, ?a)$                               & $(?q, ?q', ?\kappa)$                    & --- \\
        $\mathsf{LearnedBiKeyPose}(\cdot)$     & $(h_l, h_r, ?a)$                         & \makecell{$(?q_l, ?q_r, ?q'_l,$ \\ $?q'_r, ?\kappa)$} & --- \\
        $\mathsf{LearnedPostGrasp}(\cdot)$  & $(?h, ?o, ?p, ?a)$                       & $(?g, ?q, ?q')$                         & \makecell{$\exists e= \langle ?h,?o \rangle,$ \\$ e\in G_\mathrm{eff}$}  \\
        \bottomrule
    \end{tabular}
    \label{tab:stream_tab}
    \vspace{-0.1in}
\end{table}

\subsubsection{Stream for Policy-Trajectory Switching}

{The $\mathsf{LearnedPostGrasp}(\cdot)$ stream is designed to bridge a policy-parameterized skill $a_i$ that terminates with an in-hand object $o$ (e.g., bimanual handoff) to the next trajectory-parameterized skill  (e.g., \textit{transfer} in Appendix~\ref{sec:planned_actions}). 
The necessity for this stream arises because the precise object state and contact pose $g$, which are essential for the planning of the next skill,  are lost due to the policy's stochastic nature.
To address this, we collect the contact pose $g$ from a placement after the skill  $a_i$ in each demonstration. Thereafter, $\mathsf{LearnedPostGrasp}(\cdot)$ reuses the network $\Phi_\tau$ to predict the placement trajectory, and computes the contact pose $g$ using the procedure employed by  $\mathsf{LearnedAttach}(\cdot)$.}

\section{Plan Composition} 

This section introduces how skills are organized to solve long-horizon tasks. We first present the automated skill schema construction (Section~\ref{sec:schemata}).
Next, in Section \ref{sec:constraint_handle}, we present the implementation of physical constraints that learned skills must comply with. 
Finally, Section~\ref{sec:online} outlines the online deployment phase and the interleaved perceive-plan-act logic.

\subsection{Schemata for Learned Skills} \label{sec:schemata}

Following Section \ref{sec:streams}, the symbolic  skill schemata of traditional planned skills and learned skills are obtained differently. For planned skills in traditional TAMP literature, we hand-craft their skill schemata in our PDDL domain following the standard PDDLStream definitions of classical TAMP systems~\cite{garrett2018sampling, vu2024coast}; these primitives are summarized in  Appendix \ref{sec:planned_actions}. On the other hand, for contact-involved skills detected from demonstrations, we develop an automated workflow to build skill schemata from edge operations and network-integrated stream functions. This automated workflow leverages a small set of representative stream functions (each defined by a unique contact pattern) as reusable building blocks,
reducing the manual effort of specifying schemata for diverse skill types.

\begin{algorithm}[t!] \footnotesize
    \caption{BuildActionSchema}\label{alg BuildActionSchema}
    \LinesNumbered %
    \KwIn{Arms $\mathbf{H}$,  skill sequence $\{a_i \}^n_{i=1}$, edge operations $\{\tilde{E}_i \}^n_{i=1}$,  traditional  skill schemata $\mathfrak{A}_0$, a set of streams $\Psi$ }%
    \KwOut{Skill schema $\mathfrak{A}$ including learned skills}%
    $\mathfrak{A} \gets \mathfrak{A}_0$ \\
    \For{$i \in 1, \dots , n$}{
    \brown{\tcp{0. initialize the graph $G_0$ and skill schema $\sigma = (\text{pre,eff,args}) $ }}
    $H_i \gets \mathsf{InvolvedArms}(a_i, \mathbf{H})$ \\
    $G_0,\sigma \gets \mathsf{DefaultGraphSchema} (\rho, H_i)$ \\  \label{line:G0}
    \brown{\tcp{1. translate graphs to symbolic facts}}
    $\tilde{E}_\mathrm{pre} \gets \mathsf{GraphDiff}(a_i.G_\mathsf{pre}, G_0)$ \\ \label{line:E_pre}
    $\sigma.\mathrm{pre} \gets  \sigma.\mathrm{pre} \wedge \mathsf{EdgeOps2Predicates} (\tilde{E}_\mathrm{pre})  \wedge \mathsf{Derived}(a_i.G_\mathrm{pre})$ \\ \label{line:edge2pred}
    $\sigma.\mathrm{eff} \gets \sigma.\mathrm{eff} \wedge \mathsf{EdgeOps2Predicates} (  \tilde{E}_i ) \wedge \mathsf{Derived}(a_i.G_\mathrm{eff})$ \\  \label{line:a-eff}
    \brown{\tcp{2. associate streams with skills}}
    $ F_a \gets \mathsf{MatchStreams}(a_i, \Psi)$ \\ \label{line:matchstream}
    $\sigma.\mathrm{pre} \gets \sigma.\mathrm{pre} \wedge \mathsf{Bound}(\sigma.\mathrm{pre}, F_a.\text{domain} \wedge F_a.\text{certified} ) $\label{line:bound} \\  
    $\sigma.\mathrm{eff} \gets \sigma.\mathrm{eff}  \wedge \pddl{Applied}(a_i) $ \\ \label{line:appliedeff}
    $\sigma.\mathrm{args} \gets \mathsf{SortedArguments}(\sigma.\mathrm{pre}, \sigma.\mathrm{eff}) $ \\ \label{line:sortarg}
    $\mathfrak{A} \gets \mathfrak{A} \cup \sigma$
    }
    \Return $\mathfrak{A}$
    \end{algorithm}
    \vspace{-0.1in}

Algorithm~\ref {alg BuildActionSchema} describes the skill schemata construction for a sequence of skills $\{a_i\}^n_{i=1}$ detected by  VLM (see Section~\ref{sec:skillseg}). 
A skill schema is composed of \textit{arguments}, \textit{preconditions}, and \textit{effects}.
Here, we start with a default schema $\sigma_0$ and a default graph $G_0$ obtained from $\mathsf{DefaultGraphSchema}(\cdot)$, as in Line~\ref{line:G0}.  The creation of the default graph depends on the number of arms. An example of a default schema for the bimanual skill is:

\begin{footnotesize}
 $$
\begin{aligned}
\pddl{BimanualSkill} &(h_l, h_r, q_l, q_r, q'_l, q'_r ) \\
 \pddlkw{pre:}  & \pddl{AtConf}(h_l,q_l) \wedge \pddl{AtConf}(h_r,q_r) \\
 \pddlkw{eff:}  & \pddl{AtConf}(h_l,q'_l) \wedge \neg \pddl{AtConf}(h_l,q_l) \\
 &  \pddl{AtConf}(h_r,q'_r) \wedge \neg \pddl{AtConf}(h_r,q_r)
\end{aligned}
$$
\end{footnotesize}

The construction process can be divided into two stages, which are described in detail as follows.
\subsubsection{Stage.1 - Translating Graphs into Symbolic Predicates}
The translation from $G_\mathrm{pre}, G_\mathrm{eff}$ to the predicates is detailed in \textit{Line~\ref{line:E_pre}-\ref{line:a-eff}}, reflecting information in both paired contacts and involved entities. 
For each skill, we are interested in the geometric difference it causes, encoded in edge changes. 
In Line \ref{line:E_pre}, we obtain the set of edge changes $\tilde{E}_\mathrm{pre}$ of \textit{precondition} by comparing the graph $G_\mathrm{pre}$ and $G_0$ using $\mathsf{GraphDiff}(\cdot)$. On the other hand, the difference between $G_\mathrm{eff}$ and $G_\mathrm{pre}$ is identical to the change set $\tilde{E}_i$. 
For both \textit{precondition} and \textit{effect}, the function $\mathsf{EdgeOps2Predicates}(\cdot)$ maps edge changes to contact-related predicates using rules in Table~\ref{tab:EdgeOps2Predicates}.
Apart from the predicates reflecting pair contacts, $\mathsf{Derived}(\cdot)$ derives unary predicates according to the graph connections. For example, if one hand $h$ is not connected to any objects, a predicate $\pddl{HandEmpty}(h)$ is added. 
According to VLM-labeled object properties ($O_\mathrm{on}$, $O_\mathrm{in}$ , $O_\mathrm{movable}$, and $O_\mathrm{fixed}$), we add object-related unary predicates, such as $\pddl{Surface}$, $\pddl{Container}$, $\pddl{Movable}$, and $\pddl{Fixed}$, to further shrink the search space. 

\subsubsection{Stage.2 - Associating Streams with Skills}
Besides the translation of contact-related predicates, \textit{Line~\ref{line:matchstream}- \ref{line:bound}} links the symbolic skill to its corresponding stream function. First, the function $\mathsf{MatchStreams}(\cdot)$ assigns stream function(s) to a skill according to their contact patterns, as introduced in Section \ref{sec:streams}.  Then, $\mathsf{Bound}(\cdot)$  integrates $F_a$  by adding predicates from both its \texttt{domain} (inputs) and its \texttt{certified} (outputs) directly to the skill's precondition list. This function unifies the stream input variables with arguments already in the precondition (e.g., linking the stream's $?o$ to a known object $o_1$). It then adds the stream's certified predicates, and any free variables left over (e.g., an undecided pose $?p$) are the unbound output variables. By placing these in the precondition, the planner is forced to call the stream at runtime to compute and \textit{certify} them.
Finally, in \textit{Line \ref{line:sortarg}}, all the arguments of related predicates are collected as the parameters of the skill schema. The PDDL domain is completed by constructing a schema $\mathfrak{A}$ for each skill detected from the demonstrations.  

\subsection{Implementation of Constraint Stream Functions} \label{sec:constraint_handle} %

With discrete symbolic representation and continuous-state learning-based modeling, the  atomic skills are ready to be chained together to form a skill sequence.   
A plan skeleton is considered feasible if there exists a set of evaluated variables that satisfy all constraints.
In this section, we introduce our detailed implementation of two physical constraints, \textit{reachability} and \textit{safety}, that are enforced when chaining learned skills.

\subsubsection{Reachability Constraint}
\begin{figure}
    \centering
\includegraphics[width=0.9\linewidth]{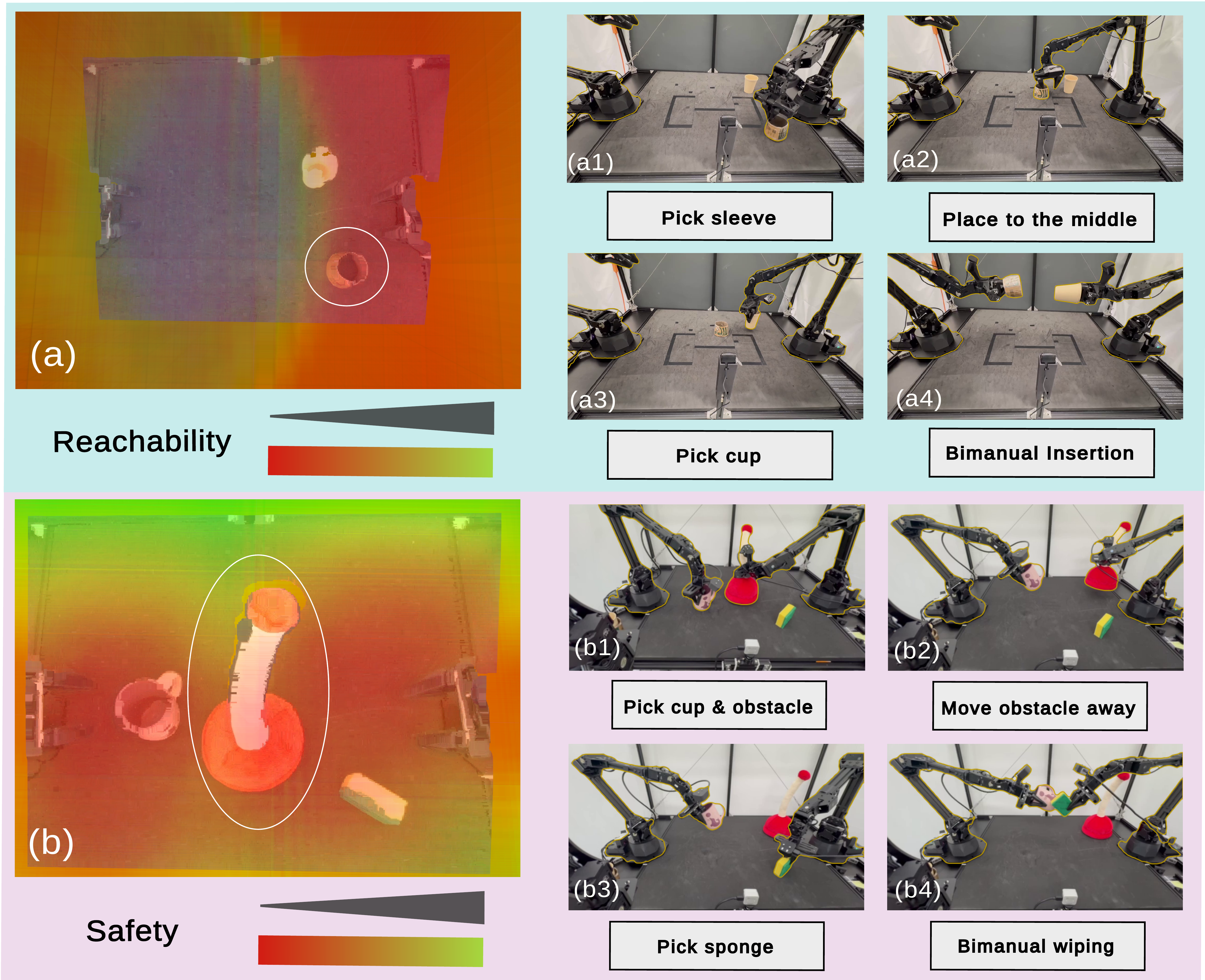}		
	\caption{ (a) The reachability map constructed for the left puppet arm of the ALOHA \cite{zhao2023learning} robot. The cup sleeve (circled) is in a low-reachability area. (a1)--(a4) are the cup-sleeve insertion task progression in the \texttt{Unreachable} setting. (b) The MDF map constructed for the cup wiping task. The toy plunger (circled) stands in an area with high collision probability.  (b1)--(b4) are the cup wiping task progression in the \texttt{Unsafe} setting. }
	\label{fig:constraint-visualize}
\vspace{-0.16in}
\end{figure}

Kinematic constraints arise frequently in bimanual tasks, so we introduce a test stream $\mathsf{IsReachable}(h, p_0)$ that checks whether the end-effector of arm $h$ can reach an object at pose $p_0$. For each arm, we precompute a compressed reachability map. We begin by sampling end-effector poses via forward kinematics from randomized joint configurations and discretizing the workspace into a fine 3D grid. For each voxel, we generate quasi-uniform orientation samples on $SO(3)$ and compute a coverage fraction: the proportion of sampled orientations within a geodesic threshold of any stored reachable orientation. This yields a 3D map that associates each voxel with its coverage fraction. Fig. \ref{fig:constraint-visualize}(a) shows the map for the left arm; lighter regions indicate higher coverage. A voxel is labeled \textit{reachable} if its coverage fraction exceeds a set threshold. At query time, $\mathsf{IsReachable}(\cdot)$ uses only the translational component of $p_0$ to retrieve the corresponding voxel’s binary reachability value. If $p_0$ is unreachable, the result is immediately returned to the task planner for alternative plans.

\subsubsection{Policy Safety Constraint}
There are cases where the deployment environment is unexpectedly cluttered, causing potential collisions in policy rollout. We opt to model the safety of a visuomotor policy as an explicit constraint using a test stream $\mathsf{IsSafePolicy}(a, \kappa)$,  certifying whether the skill $a$ with the policy identifier $\kappa$ is safe from collision with other objects at that moment. 
Due to the stochastic nature of a learned policy, it is intractable to check its safety by forward simulating its execution trajectory.
Hence, conventional collision-checking streams such as $\mathsf{IsSafeTraj}(\cdot)$ are incompatible with visuomotor policies, which motivates our test stream implementation tailored for visuomotor policies.  

Following the practice of a recent multi-arm TAMP system~\cite{garrett2025schedulestream} in deciding arm-arm collision, we adopt a swept-volume strategy.
In detail, we construct a distance map  to approximate potentially risky regions for each policy-parameterized skill.
Denote the timespan during the contact-rich segment of skill $a$  as  $T_a$.  We replay the demonstration during  $T_a$ in simulation. At each timestep $t \in T_a$, we voxelize the workspace and record the voxels occupied by the manipulator or related objects as an occupancy map $\mathfrak{O}_t$. We then compute the Euclidean distance transform (EDT) map \cite{chen2022gpu} of $\mathfrak{O}_t$, which assigns every voxel its distance to the nearest occupied voxel. A minimum distance field (MDF) is then obtained by aggregating across timesteps, updating each voxel in the workspace with the minimum distance value encountered. The distance value of a voxel $v$ in an MDF is defined as
\begin{gather}
\textit{MDF}(v) = \min_{t \in T_a} \ \min_{u \in \mathfrak{O}_t} \ ||u - v||_2 ,
\end{gather}
where $u$ ranges over the occupied voxels in $\mathfrak{O}_t$, so the inner minimization is the per-timestep EDT and the outer one aggregates these maps into the MDF.
An example can be visualized in Fig.~\ref{fig:constraint-visualize}(b), where darker voxels mean larger collision probability.  
During the execution phase, the occupied voxels of unexpected obstacles can be queried efficiently, and any voxel with a low distance value has a high likelihood of colliding during the rollout of the visuomotor policy. Our method enables the planner to detect any unknown blocking object in an open world, without the knowledge   obstacle geometry (as required in  \cite{xue2025demogen,pan2025one}) or privileged information (e.g., obstacle 6D pose required in  \cite{jung2025rail}).
As the MDF is built offline from the swept volumes of demonstrations, it may not be accurate when the imitation policy drifts. We also observe that the MDF check is conservative: among 57 rejected obstacle poses, 17 (about 30\%) lie near the boundary of the 3\,cm margin and would execute without collision.

At the symbolic level, a  derived predicate $\mathsf{IsSafePolicy}$ is certified  when all objects on the table are in the safety area of an MDF map
:
\begin{quote}\small\ttfamily
    (:derived (IsSafePolicy ?a ?$\kappa$)\\
    \hspace*{1em}(forall (?o ?p) (imply\\
    \hspace*{2em}(and (AtPose ?o ?p) (Movable ?o) )\\
    \hspace*{2em}(CFreeMDF ?o ?p ?a ?$\kappa$))))
    \end{quote}

By design, a falsified $\mathsf{IsSafePolicy}$ predicate identifies the obstacle $o$ at pose $p$ that obstructs the motion. This geometric information is passed back to the symbolic layer, so the planner can relocate the obstacle. By contrast, solvers whose geometric tests are confined to the low-level binding stage, e.g., IDTMP~\cite{dantam2018incremental}, TAMPER~\cite{pan2024task}, and COAST~\cite{vu2024coast}, discard geometric information through their failure-to-constraint feedback.

\subsubsection{Real-world Constraint Handling}
In Fig.~\ref{fig:constraint-visualize}, two real-world experiments illustrate the constraint handling procedure. The \texttt{unreachable} condition is set up for the \textit{Cup-sleeve Insertion} task, and the \texttt{unsafe} condition for the \textit{Cup Wiping} task.
In the \texttt{unreachable} case (upper row), the cup sleeve starts outside the left arm's reachability map. The precondition of a pick skill,  $\mathsf{IsReachable}$, fails for the sleeve at its pose.  The failed check carries the unreachable object and its pose as feedback to the symbolic layer, informing the TAMP solver that the pick is not applicable unless changing the sleeve's $\mathsf{AtPose}$.  \ours therefore inserts an extra \textit{regrasp} that transports the sleeve to a reachable region, after which the cup is picked and the bimanual insertion proceeds.
In the \texttt{unsafe} case (lower row), a toy plunger is placed at the table center as an unexpected obstacle. Its occupied voxels lie outside the safe region of $\mathrm{MDF}_{\kappa}$, so the per-object check $\mathsf{CFreeMDF}(o,p,a,\kappa)$ fails for the plunger. This falsifies the derived predicate $\mathsf{IsSafePolicy}(a,\kappa)$, which is a precondition of the bimanual wiping skill $a$.   According to the definition of derived predicate  $\mathsf{IsSafePolicy}$,  the TAMP solver can satisfy it only by changing the  pose $p$ of the movable plunger $o$. To that end, \ours plans an extra pick-and-place movement, grasping the plunger with a general-purpose grasp generator~\cite{gualtieri2016high} and places it at a safe pose.

\begin{algorithm}[t!] \footnotesize
\caption{Online Planning and Adaptation}\label{alg svipexec}
\LinesNumbered
\KwIn{Skill schemata $\mathfrak{A}$, goal set $\eta$, arms $\mathbf{H}$, objects $\mathbf{O}$}

$s_0 \gets \mathsf{Perceive}()$ \label{line:perceive} \\
\brown{\tcp{Line \ref{line.TeleportTAMP}: compute a coarse skill-level skeleton}}
$\vec{a}_{\mathrm{glb}} \gets \mathsf{TeleportTAMP}(s_0, \eta, \mathfrak{A})$ \label{line.TeleportTAMP} \\

$i \gets 0$ \\

\While{$i <  |\vec{a}_{\mathrm{glb}}|$}
{
    $a_i \gets \vec{a}_{\mathrm{glb}}[i]$ \\

    $s_i \gets \mathsf{Perceive}()$ \label{line:online-perceive} \\
        \brown{\tcp{Line \ref{line:detailed-tamp}: solve a detailed TAMP problem for the current subgoal}}
    $\tau_i \gets \mathsf{DetailedTAMP}(s_i, \mathsf{Subgoal}(a_i), \mathfrak{A})$ \label{line:detailed-tamp} \\

    \brown{\tcp{Line \ref{line:execute-verify}: execute the commands and verify their effects}}
    $\mathrm{success} \gets \mathsf{ExecuteAndVerify}(\tau_i)$ \label{line:execute-verify} \\

    \If{$\mathrm{success}$}
    {
        $i \gets i + 1$ \label{line:advance-skeleton} \\
    }
}
\end{algorithm}

\subsection{Online Execution, Perception and Adaptation} \label{sec:online} %
Traditional TAMP often executes a complete motion-level plan in an open-loop manner, assuming that the world evolves exactly as predicted by the plan.
However, the object end pose after placement may drift due to unmodeled friction and sliding, and the achieved grasp pose may differ from the nominal one because the object can shift, rotate, or bounce during interaction.
Moreover, grasps may fail because of perception or execution noise.
These discrepancies accumulate across a manipulation sequence, reducing the reliability of a multi-stage plan.

\begin{figure}
    \centering
    \includegraphics[width=0.9\linewidth]{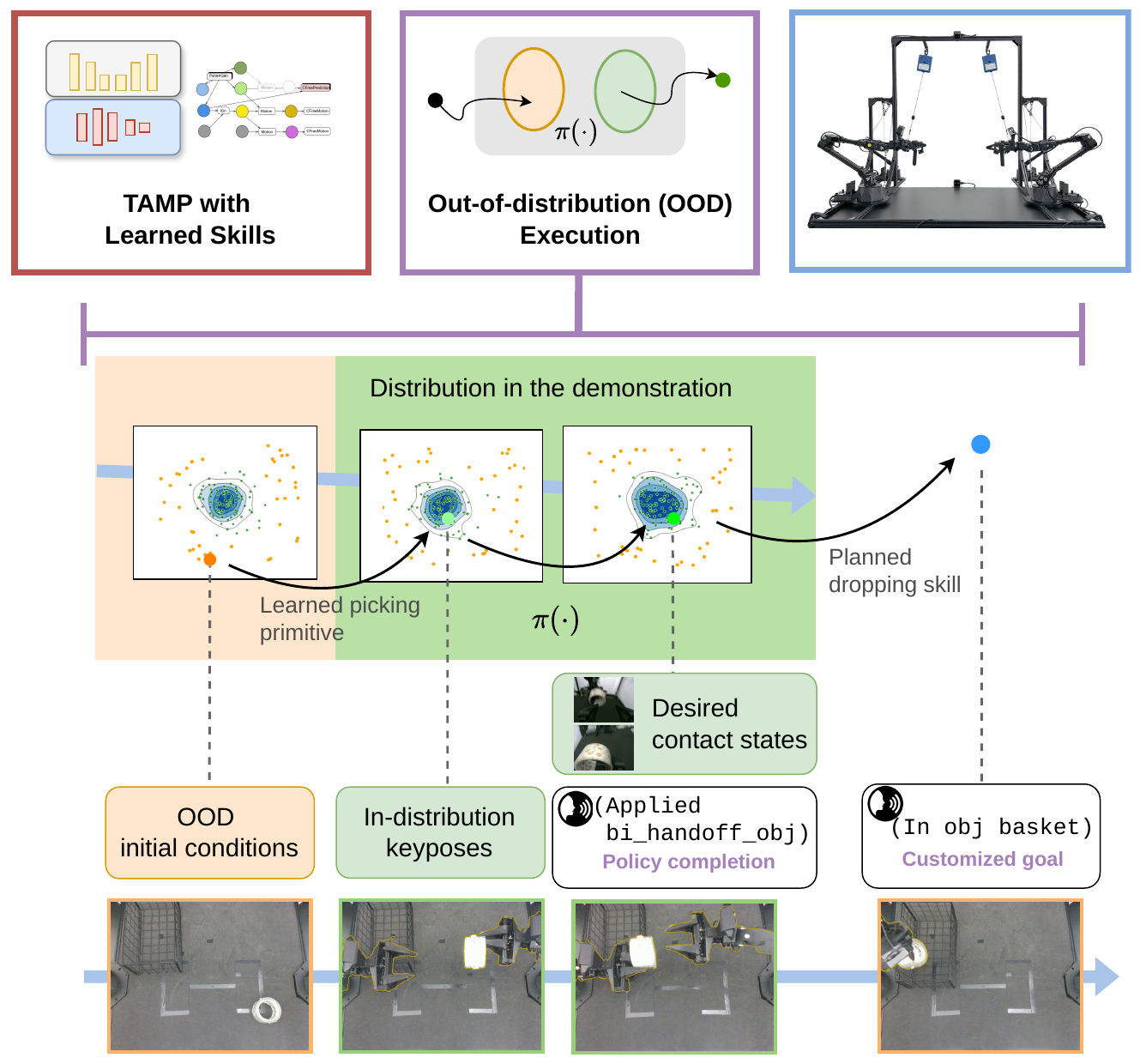}
    \caption{ An execution example for the task ``\textit{(1) Apply the handoff skill; (2) Place the object inside the basket.}''  When encountering unseen setups, \textit{\ours} leverages the learned object-picking skill to initiate the visuomotor policy in an in-distribution observation. Additionally, \textit{\ours} can complete test-time customized goals with skills in its repertoire.  }
    \label{fig:ood-2-id}
    \vspace{-0.2in}
\end{figure}

To address this issue, we interleave execution with perception and local planning.
Algorithm~\ref{alg svipexec} presents our online planning and adaptation workflow.
The system first computes a coarse skill-level skeleton and then executes it through a local \textit{perceive-plan-act-verify} loop.
The input goal $\eta$ is typically defined as applying all skills of interest, i.e., $\eta= \bigwedge^{n}_{i=1} \texttt{Applied}(a_i)$, while also allowing human preferences to be incorporated, as in Section~\ref{sec:real-exp}.
The function $\mathsf{Perceive}(\cdot)$ (Line~\ref{line:perceive}) obtains an initial state $s_0$ containing predicates of arms and task-related objects, following the implementation of M0M~\cite{curtis2022long}.

With the constructed PDDL problem $(s_0, \eta)$ and the skill schemata $\mathfrak{A}$, $\mathsf{TeleportTAMP}(\cdot)$ uses a focused TAMP solver in a coarse teleport mode to compute a global skill-level skeleton $\vec{a}_\mathrm{glb}$ (Line~\ref{line.TeleportTAMP}).
The coarse skeleton specifies the sequence of high-level actions to be achieved, including object-centric primitive actions and visuomotor policy actions, while abstracting away detailed transit and transfer motions.
For policy-parameterized actions, the skeleton also carries the policy identifier $\kappa$, which determines the policy rollout and its contact-based termination.

Online execution then proceeds with an index $i$ over the global skeleton.
At each iteration, the system first refreshes the current state estimate by calling $\mathsf{Perceive}(\cdot)$ (Line~\ref{line:online-perceive}).
Given the current high-level action $a_i$, we extract its action-level subgoal $\mathsf{Subgoal}(a_i)$ and call $\mathsf{DetailedTAMP}(\cdot)$ to solve a local TAMP problem from the refreshed state $s_i$ to this subgoal (Line~\ref{line:detailed-tamp}).
This local planner materializes the current abstract action by regenerating the necessary learned stream outputs, connector motions, primitive trajectories, or synchronized commands that invoke the visuomotor policy identified by $\kappa$.

The resulting command sequence $\tau_i$ is executed and checked by $\mathsf{ExecuteAndVerify}(\cdot)$ (Line~\ref{line:execute-verify}).
Planned trajectories are streamed to the robot, while learned primitives and visuomotor policies are executed by their corresponding low-level controllers.
To monitor completion of learned primitives and policies, we additionally train a \textit{contact detector} that detects the gripper-object contact status, i.e., $\tilde{E}_{h-o}$, which has already been labeled in Sec.~\ref{sec:skillseg}.  The contact detector is a multi-task classifier, where ResNet-18 encodes the wrist-camera images and outputs a binary value conditioned on a gripper-object edge $\langle h, o \rangle$ (e.g., \texttt{right\_arm\_cup}). In addition to the data obtained from demonstrations, we record five self-play episodes in which the arm moves randomly with the gripper closed, thus preventing false-positive detections when a grasp misses. During online rollout, \ours terminates a learned policy when the desired contact effect is observed, or triggers replanning when a desired contact is not established.
The reliability of the contact detector under cluttered backgrounds and novel instances is evaluated in Sec.~\ref{sec:stress-tests}.

If the verified effect of $a_i$ is achieved, the system advances to the next high-level action by setting $i \gets i+1$ (Line~\ref{line:advance-skeleton}).
Otherwise, the index is not advanced.
The next iteration therefore re-perceives the scene and invokes $\mathsf{DetailedTAMP}(\cdot)$ again for the same action-level subgoal.

In summary, the global skill skeleton provides long-horizon structure, while each local TAMP problem is repeatedly grounded in the latest observed state, discarding stale motion details.

\section{Experiments and Results} %
In this section, we evaluate \ours in both simulated tasks and  real-world tasks, providing empirical results to answer the following questions: 
\begin{itemize}
    \item \textbf{Q1}: Can \ours perform effectively in previously unseen tabletop setups? 
    \item \textbf{Q2}: Can \ours manage reachability and collision-free constraints when involving learned policies?
    \item \textbf{Q3}: Can \ours complete tasks with novel goals that require multiple steps?
    \item \textbf{Q4}:  Can \ours outperform variants that use a single method (e.g., image-based policy, equivariant policy, or traditional TAMP)?
\end{itemize}

\subsection{Spatial Generalization}
To answer \textbf{Q1}, we carry out both simulated and real-world experiments on the ALOHA \cite{zhao2023learning} platform.  We evaluate the performance of \ours when it encounters both in-distribution (\texttt{ID}) and out-of-distribution (\texttt{OOD}) observations during inference. 

\subsubsection{Simulation Evaluations}  \label{sec:sim}  \label{sec:sim_setup}
We use the peg-in-hole environment from ALOHA~\cite{zhao2024aloha}, where an ALOHA robot has to insert a peg into a socket. We set three object placement distributions:
\begin{itemize}
\item \texttt{ID}: each object’s 2D position is uniformly sampled within a $0.2,\text{m} \times 0.2,\text{m}$ region with fixed orientation, mirroring that in ALOHA~\cite{zhao2024aloha}.
\item \texttt{XY-OOD}: expands sampling to a $0.25,\text{m} \times 0.25,\text{m}$ region (superset of \texttt{ID}), generating position-shifted \texttt{OOD} observations.
\item \texttt{XYH-OOD}: further randomizes orientation by sampling the heading $\omega \in (-0.5\pi, 0.5\pi)$ on top of the \texttt{XY-OOD} positional variance, producing more challenging \texttt{OOD} observations.
\end{itemize}

We only use 50 demonstration episodes collected in \texttt{ID} for training, while all three placement distributions are used for testing.
We compare \ours against ACT \cite{zhao2023learning} and DP \cite{chi2023diffusion} in Table~\ref{tab:sim_ood}.
ACT and DP often fail during insertion due to sensitivity to contact poses, even in the \texttt{ID} setting.
Under \texttt{XYH-OOD}, \ours achieves 88\% success, compared with 12\% for the stronger baseline DP.
Although \ours has a slightly longer average trial duration, this is largely because ACT and DP often fail early before reaching the \texttt{OOD} objects.

\begin{table}
    \centering
        \caption{Success rates and completion times of different methods in the simulated \textit{Peg-in-Hole} task. Brackets show 95\% Wilson confidence intervals over 50 trials.}
    \begin{tabular}{l l c c}
    \toprule
         \textbf{Setup} & \textbf{Method} & \makecell[c]{\textbf{Success}\\\textbf{Rate (\%)}$\uparrow$} & \makecell[c]{\textbf{Completion}\\\textbf{Time (s)}$\downarrow$}  \\
     \midrule
       \texttt{ID}      & ACT \cite{zhao2023learning} & 44 {\scriptsize[31, 58]} & 7.52  $\pm$ 1.32 \\
                        & DP \cite{chi2023diffusion}  & 54 {\scriptsize[40, 67]} & 8.66 $\pm$ 1.14 \\
                        & \ours                      & \textbf{100} {\scriptsize[93, 100]} & 10.21 $\pm$ 1.84 \\
     \midrule
        \texttt{XY-OOD}  & ACT \cite{zhao2023learning} & 38 {\scriptsize[26, 52]} & 7.98 $\pm$ 1.29 \\
                         & DP \cite{chi2023diffusion}  & 42 {\scriptsize[29, 56]} & 9.22 $\pm$ 0.97 \\
                         & \ours                     & \textbf{100} {\scriptsize[93, 100]} & 10.75 $\pm$ 2.30 \\
     \midrule
        \texttt{XYH-OOD} & ACT \cite{zhao2023learning} & 2 {\scriptsize[0, 10]} & -- \\
                         & DP \cite{chi2023diffusion}  & 12 {\scriptsize[6, 24]} & 9.44  $\pm$ 1.02 \\
                         & \ours                      & \textbf{88} {\scriptsize[76, 94]} & 12.75 $\pm$ 2.46 \\
    \bottomrule
    \end{tabular}
    \label{tab:sim_ood}
    \vspace{-0.1in}
\end{table}

\subsubsection{Real-world Evaluations} \label{sec:real-exp}
We conducted real-world experiments on the ALOHA \cite{zhao2023learning} platform with an overhead RGB-D camera for 3D perception.
Four real-world tasks are designed for evaluation: (T1) \textit{Object Handoff}, which is successful when the object is firmly grasped by the left gripper after being handed off from the right gripper; (T2) \textit{Screwdriver Packing}, which is successful when the screwdriver is placed inside its case with the lid fully closed; (T3) \textit{Cup-sleeve Insertion}, which is successful when a paper cup is fully inserted into a cup sleeve; (T4) \textit{Cup-Sponge-Screwdriver}, which composes the screwdriver packing, cup handoff, and cup wiping skills (detailed in Sec.~\ref{sec:res_combine_skills}).
All four tasks are evaluated under \texttt{ID} and \texttt{OOD} settings, with object placements sampled from larger regions under \texttt{OOD}. Tasks T1--T3 additionally include a \texttt{Leaky} setting, where target objects are tilted against static supports. Detailed placement settings are available on the project website.

\begin{table}[t]
    \centering
    \caption{Success rates and completion times  in real-world tasks under different initial setups. Statistics obtained from 20 trials. Brackets show 95\% Wilson confidence intervals.}
    \setlength{\tabcolsep}{2pt}
    \begin{tabular}{@{}L{0.18\columnwidth}L{0.11\columnwidth}L{0.16\columnwidth}C{0.24\columnwidth}C{0.22\columnwidth}@{}}
    \toprule
       \textbf{Task} & \textbf{Setup} & \textbf{Method} & \makecell[c]{\textbf{Success}\\\textbf{Rate (\%)}$\uparrow$} & \makecell[c]{\textbf{Completion}\\\textbf{Time (s)}$\downarrow$} \\ \midrule
\multirow{9}{*}{\makecell[l]{Object\\Handoff}}
& \texttt{ID} & DP \cite{chi2023diffusion} & 90 {\scriptsize[70, 97]} & 19.82 $\pm$ 0.88 \\
&             & $\pi_{0.5}$                & \textbf{100} {\scriptsize[84, 100]} &  15.60 $\pm$ 0.33 \\
&             & \ours                     & \textbf{100} {\scriptsize[84, 100]} & 21.09 $\pm$ 1.07 \\ \cmidrule(lr){2-5}
& \texttt{OOD} & DP \cite{chi2023diffusion} & 30 {\scriptsize[15, 52]} & 20.98  $\pm$ 0.78  \\
&              & $\pi_{0.5}$                & 20 {\scriptsize[8, 42]} &  15.95 $\pm$ 0.45 \\
&              & \ours                     & \textbf{95} {\scriptsize[76, 99]}  & 21.78 $\pm$ 1.18 \\  \cmidrule(lr){2-5}
& \texttt{Leaky} & DP \cite{chi2023diffusion} & 60 {\scriptsize[39, 78]} & 20.67 $\pm$ 0.65 \\
&                & $\pi_{0.5}$                & 90 {\scriptsize[70, 97]} &  17.60 $\pm$ 0.50 \\
&                & \ours                     & \textbf{100} {\scriptsize[84, 100]} & 22.43  $\pm$ 1.82 \\ \midrule
\multirow{9}{*}{\makecell[l]{Screwdriver\\Packing}}
& \texttt{ID} & DP \cite{chi2023diffusion} & 55 {\scriptsize[34, 74]} & 33.65 $\pm$ 1.24 \\
&             & $\pi_{0.5}$                & 60 {\scriptsize[39, 78]} &  28.64 $\pm$ 1.12 \\
&             & \ours                     & \textbf{80} {\scriptsize[58, 92]}  & 34.89 $\pm$ 0.98 \\ \cmidrule(lr){2-5}
& \texttt{OOD} & DP \cite{chi2023diffusion} & 10 {\scriptsize[3, 30]} & -- \\
&              & $\pi_{0.5}$                & 10 {\scriptsize[3, 30]} & -- \\
&              & \ours                     & \textbf{80} {\scriptsize[58, 92]}  & 36.34 $\pm$ 3.44 \\ \cmidrule(lr){2-5}
& \texttt{Leaky} & DP \cite{chi2023diffusion} & 0 {\scriptsize[0, 16]} & -- \\
&                & $\pi_{0.5}$                & \textbf{75} {\scriptsize[53, 89]} &  30.50 $\pm$ 1.49 \\
&                & \ours                     & 65 {\scriptsize[43, 82]}  & 35.21 $\pm$ 2.34 \\ \midrule
\multirow{9}{*}{\makecell[l]{Cup-sleeve\\Insertion}}
& \texttt{ID} & DP \cite{chi2023diffusion} & 75 {\scriptsize[53, 89]} & 17.64  $\pm$ 0.99 \\
&             & $\pi_{0.5}$                & \textbf{85} {\scriptsize[64, 95]} &  15.42 $\pm$ 0.49  \\
&             & \ours                     & 80 {\scriptsize[58, 92]}  & 20.73 $\pm$ 1.32  \\ \cmidrule(lr){2-5}
& \texttt{OOD} & DP \cite{chi2023diffusion} & 35 {\scriptsize[18, 57]} & 19.21 $\pm$ 1.44 \\
&              & $\pi_{0.5}$                & 25 {\scriptsize[11, 47]} &  16.80 $\pm$ 0.67 \\
&              & \ours                     & \textbf{75} {\scriptsize[53, 89]}  & 21.20 $\pm$ 3.34 \\ \cmidrule(lr){2-5}
& \texttt{Leaky} & DP \cite{chi2023diffusion} & 35 {\scriptsize[18, 57]} & 18.54 $\pm$ 1.39 \\
&                & $\pi_{0.5}$                & 35 {\scriptsize[18, 57]} &  15.03 $\pm$ 2.23 \\
&                & \ours                     & \textbf{40} {\scriptsize[22, 61]}  & 20.39 $\pm$ 3.57 \\ \midrule
\multirow{6}{*}{\makecell[l]{Cup-Sponge-\\Screwdriver}}
& \texttt{ID}  & DP \cite{chi2023diffusion} & 20 {\scriptsize[8, 42]} & 56.98 $\pm$ 4.25  \\
&              & $\pi_{0.5}$                & \textbf{70} {\scriptsize[48, 85]} & 54.27 $\pm$ 4.72  \\
&              & \ours                      & 55 {\scriptsize[34, 74]} &  55.70 $\pm$ 5.72 \\ \cmidrule(lr){2-5}
& \texttt{OOD} & DP \cite{chi2023diffusion} & 0 {\scriptsize[0, 16]} & -- \\
&              & $\pi_{0.5}$                & 10 {\scriptsize[3, 30]} &  54.45 $\pm$ 4.90  \\
&              & \ours                      & \textbf{50} {\scriptsize[30, 70]} & 58.02 $\pm$ 5.66  \\
    \bottomrule
    \end{tabular}
    \label{tab:real_ood}
    \vspace{-0.1in}
\end{table}

In the real-world experiments, we use DP and $\pi_{0.5}$, a stronger pretrained behavior model, as baselines. For each task,  20 demonstrations are collected (60 for \textit{Cup-Sponge-Screwdriver}, 20 per skill) to train DP and \ours from scratch, and the same data are used to fine-tune $\pi_{0.5}$.
 Table~\ref{tab:real_ood} reports success rates and completion time averaged over 20 rollouts per task per setting.

Results show that $\pi_{0.5}$ is competitive with \ours under \texttt{ID} and \texttt{Leaky} placements, while \ours outperforms both DP and $\pi_{0.5}$ under \texttt{OOD} placements.
The same pattern holds in the long-horizon \textit{Cup-Sponge-Screwdriver} task, where DP fails under all \texttt{OOD} placements.
In \textit{Cup-sleeve Insertion}, success under \texttt{Leaky} is low for all methods due to the fine-grained requirement and distribution shift. Noisy observations from single-view depth sensing also cause the learned primitives in \ours to drift.

\subsection{Constraint Handling}

This section aims to address the core concern of \textbf{Q2} by showing the capability of \ours to replan additional movements to mitigate the constraint violations. 
To demonstrate that, we design two simulation scenarios based on the peg-in-hole task, namely {\texttt{unreachable}} and {\texttt{unsafe}}. 
These two scenarios are intended to evaluate the proposed constraints $\mathsf{IsReachable}(\cdot)$ and $\mathsf{IsSafePolicy}(\cdot)$, respectively, as described in Section~\ref{sec:constraint_handle}.
Real-world demonstrations of these two cases were shown earlier in Section~\ref{sec:constraint_handle} (Fig.~\ref{fig:constraint-visualize}).

\begin{figure}[htb]
    \centering
    \includegraphics[width=\linewidth]{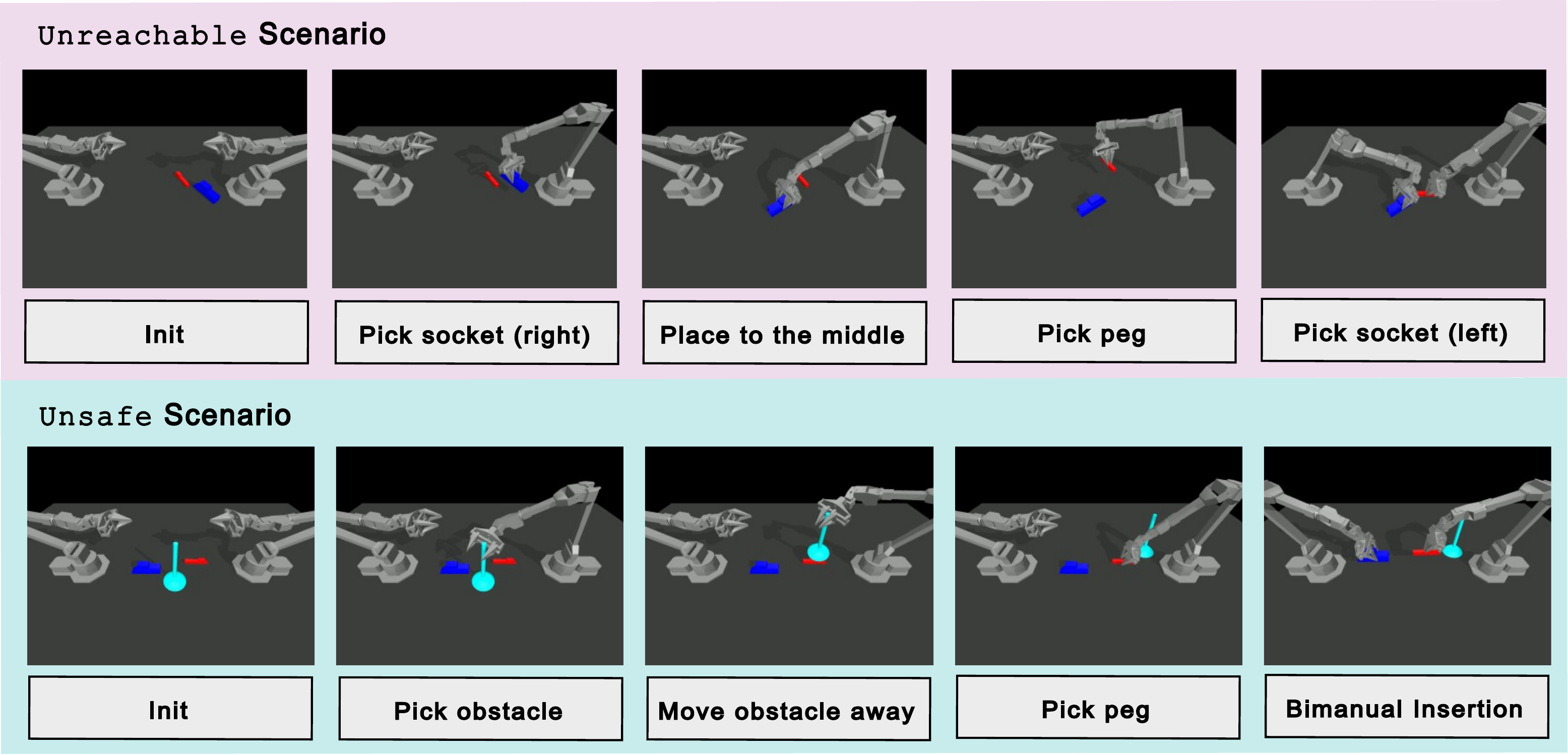}
    \caption{Executions in \texttt{unreachable} and \texttt{unsafe} scenarios.}
    \label{fig:unsafe-unreach}
\end{figure}

As shown in Figure~\ref{fig:unsafe-unreach}, the initial object placement in the \texttt{unreachable} scenario disables either one of the robot arms from reaching the target object,   testing  \ours ability to handle reachability constraints.
In the \texttt{unsafe} scenario, a pole is introduced as a potential obstruction, testing the system’s ability to cope with unexpected obstacles.

\begin{table}
    \centering
    \caption{Results of \ours's ability to cope with unsatisfying motion constraints. Statistics averaged across 20 trials. Each trial is capped at 60 seconds and timeouts are counted as failures. 
    }
    \begin{tabular}{c c c}
    \toprule
         \textbf{Setup} &   \makecell[c]{\textbf{Success}\\\textbf{Rate (\%)}$\uparrow$} &  \makecell[c]{\textbf{Initial Planning}\\\textbf{Time (s)}$\downarrow$}  \\
         \midrule
         \texttt{ID} &  100 &  7.55 $\pm$ 0.35 \\
       \texttt{unreachable}     & 55 & 12.36 $\pm$ 1.51 \\
    \texttt{unsafe}  &   70 & 27.39 $\pm$ 2.36 \\
    \bottomrule
    \end{tabular}
    \label{tab:sim_long}
    \vspace{-0.1in}
\end{table}

Table~\ref{tab:sim_long} reports success rates and initial planning time, i.e., the time to find the first feasible plan.
Baseline methods ACT \cite{zhao2023learning} and DP \cite{chi2023diffusion} are omitted because they fail entirely in both \texttt{unreachable} and \texttt{unsafe} scenarios.
As shown in Fig.~\ref{fig:unsafe-unreach}, in the \texttt{unreachable} scenario, the robot uses the right arm to transport the socket from  the right half of the table  to the middle of the table, which is accessible by the left arm.
In the \texttt{unsafe} scenario, the robot first repositions the obstructing pole to ensure a collision-free bimanual insertion operation. Compared to \texttt{ID}, success rates drop primarily due to unexpected collisions caused by noisy or incomplete point-cloud observations, while initial planning time increases with the search space. Further discussion is provided in Section~\ref{sec:discussion}.

\subsection{Evaluation in  Skill Composition and Reusing} \label{sec:res_combine_skills}

\begin{figure}[t!]
    \centering
    \includegraphics[width=1\linewidth]{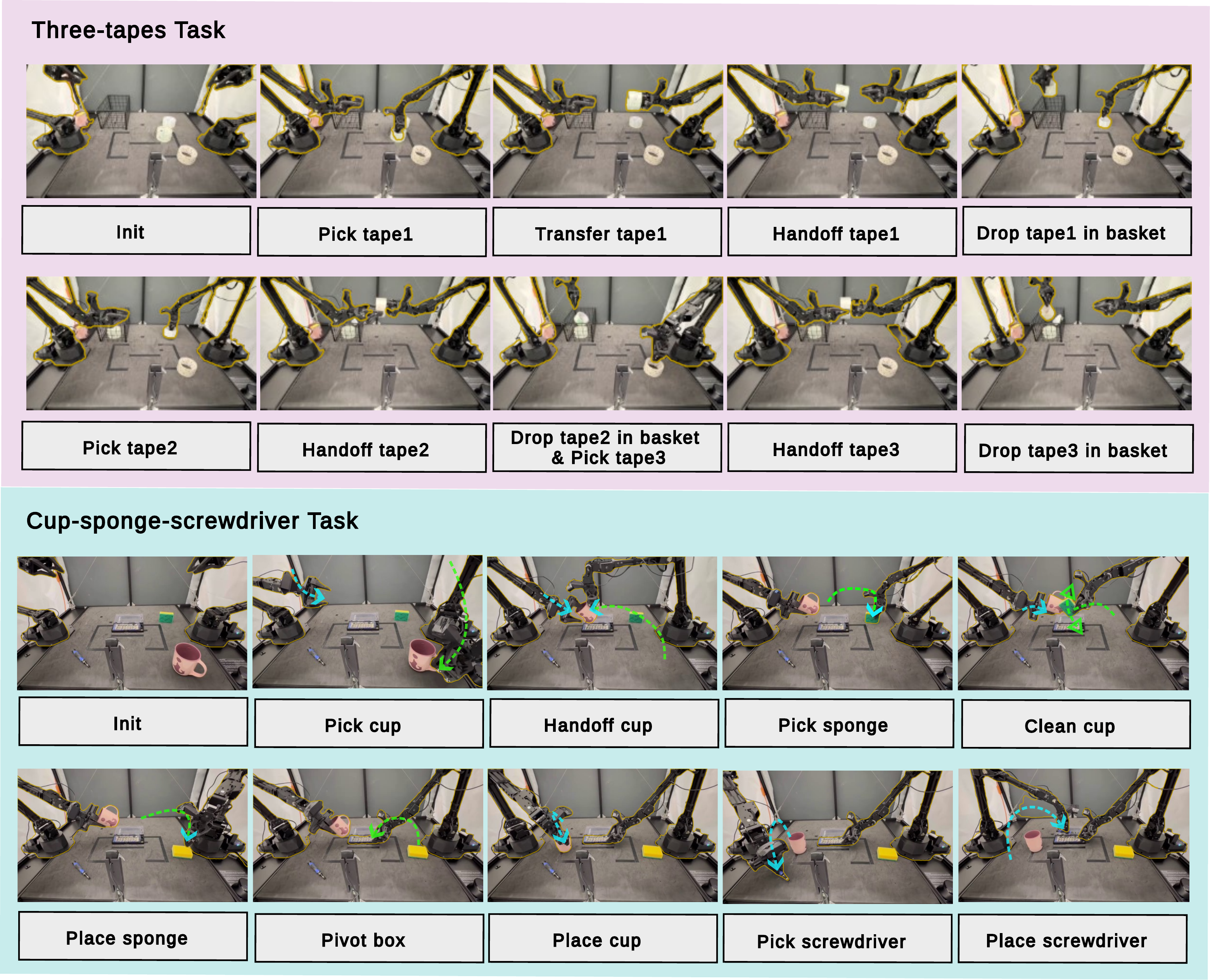}
    \caption{Experiments of two long-horizon tasks with customized goals.}
    \label{fig:long-tasks}
    \vspace{-0.1in}
\end{figure}

The experiments in this section demonstrate that \ours can compose skills from its repertoire to solve long-horizon, human-specified goals, addressing \textbf{Q3}. Besides the skill-termination predicates (e.g., \pddl{Applied}($\cdot$)), customized goal preference (e.g., \pddl{On}($\cdot$), \pddl{In}($\cdot$)) is also incorporated.
We evaluate two representative long-horizon goals:
\begin{itemize}
    \item \textit{Three-Tapes}: Using only the skill learned for \textit{Object Handoff}, \ours completes a repeated-skill objective
    $\eta_1 = \forall o \in O_\mathrm{tapes}, \pddl{In}(o, o_\mathrm{basket})$,
    where the basket $o_\mathrm{basket}$ is fixed at the top-left of the table.  Notably, the goal does not specify which skill to execute, requiring the system to autonomously select skills.

    \item \textit{Cup-Sponge-Screwdriver}: Given three distinct sets of demonstrations on screwdriver packing skill $a_1$, a cup handoff skill $a_2$, and a cup wiping skill $a_3$, \ours is able to complete the task goal  $\eta_2 =   \{ \forall i \in \{1,2,3\}, \pddl{Applied}(a_i) \wedge \{ \forall o \in O_\mathrm{movable},  \pddl{On} (o, o_\mathrm{table}) \} \}$.  Success requires handing the cup from the right to the left gripper, contacting it with the sponge held by the right gripper, packing the screwdriver with the lid closed, and leaving all objects on the table.
\end{itemize}
 Same as before, \ours performs skill decomposition described in Section \ref{sec:skillseg}, and trains each contact-rich segment using DP \cite{chi2023diffusion} with 20 episodes of demonstrations.
Following DexMimicGen~\cite{jiang2025dexmimicgen}, we implement a queue-like scheme enabling parallel execution of the planned and learned trajectories, reducing overall runtime.
Fig. \ref{fig:long-tasks} shows screenshots of these two complex long-horizon tasks. 
Counting skills including \textit{transit} and \textit{transfer}, the \textit{Three-Tapes} task is completed by a 21-step plan skeleton featuring a repeated bimanual handoff skill, while the \textit{Cup-Sponge-Screwdriver} task uses a 19-step skeleton with three distinct bimanual operations. 

We also evaluate the effect of improving an individual skill in the \textit{Cup-Sponge-Screwdriver} task.
The three original skills are trained with all task objects on the tabletop. In addition, we train a clean-background variant of the screwdriver packing skill $a_1$, denoted by $a_4$, with the sponge and cup removed. When substituting $a_4$ for the original cluttered-background skill ($a_1$) in goal $\eta_2$, execution suffers from unnecessary idle states during screwdriver insertion,  which is likely due to increased uncertainty from noisy observations caused by the distracting objects.  On the other hand, using the skill $a_1$ trained with all objects in the background,
both the success rate and the average number of completed steps rise (from 35\% and 10.35 steps with the clean-background variant $a_4$ to 50\% and 15.30 steps with the cluttered-background skill $a_1$).

\subsection{Comparative and Ablation Studies} \label{sec:comparison}
To demonstrate the rationale and advantages of the design choices adopted in \ours, we carry out additional comparative and ablation studies on LIBERO \cite{liu2023libero}, DexMimicGen \cite{jiang2025dexmimicgen}, {and real-world tasks}. 

\begin{table}[ht]
\centering
\caption{Success rates in LIBERO dataset evaluation. }
\footnotesize
\setlength{\tabcolsep}{1pt}
\begin{tabular}{@{}L{0.34\columnwidth}C{0.11\columnwidth}C{0.11\columnwidth}C{0.12\columnwidth}C{0.12\columnwidth}C{0.12\columnwidth}@{}}
\toprule
\textbf{Method} & \textbf{Spatial} & \textbf{Object} & \textbf{Long1} & \textbf{Long5} & \textbf{Long8} \\
\midrule
DP \cite{chi2023diffusion} (10 demos per task)  & 66\% & 58\% & 20\% & 50\% & 10\% \\
DP \cite{chi2023diffusion} (50 demos per task) & 84\% & 78\% & 30\% & \textbf{90}\% & 70\% \\
NOD-TAMP \cite{cheng2023nod} & 84\% & 94\% & 70\% & 70\% & \textbf{90}\% \\
\ours (Open-loop GPD)  & 34\% & 15\% & 0\% & 0\% & 0\% \\
\ours (Open-loop M2T2)  & 82\% & 53\% & 0\% & 20\% & 10\% \\
\ours (Open-loop) & 95\% & 96\% & 80\% & 90\% & 30\% \\
\rowcolor{gray!15} \ours (Closed-loop) & \textbf{98}\% & \textbf{98}\% & \textbf{80}\% & \textbf{90}\% & \textbf{90}\% \\
\bottomrule
\end{tabular}
\vspace{-0.1in}
\label{tab:libero_results}
\end{table}

\paragraph{Evaluation of Object-centric Primitives}
To isolate the performance of object-centric primitives, we select prehensile tasks from LIBERO \cite{liu2023libero} for evaluation: the \textbf{Spatial} and \textbf{Object} sets (each with 10 tasks) plus three tasks from the \textbf{Long} set. 

We compare against 1)  Two DP \cite{chi2023diffusion} variants trained on 10 and 50 demonstrations per task; 2) NOD-TAMP \cite{cheng2023nod}, with results on the same task subsets taken from their paper since the code is not open-source;   3) An open-loop ablation of \ours without online adaptation; and 4) TAMP-inspired baselines, \ours (Open-loop GPD) and \ours (Open-loop M2T2). In detail, we replace the learned primitives with generic grasp planners, GPD \cite{ten2017grasp} and M2T2 \cite{yuan2023m2t2}, that are inspired by the grasp-grounding modules of TAMPER \cite{pan2024task} and TiPToP \cite{shen2026tiptop}, respectively.

Table~\ref{tab:libero_results} presents the success rates averaged on 10 rollouts per task.
\ours (closed-loop) matches or outperforms all baselines across the five suites. On \textbf{Long8}, its success rate rises from 30\% in the open-loop variant to 90\% with online perception, consistent with errors from stale object estimates and collisions during open-loop execution.

\begin{figure}[t]
\centering
\includegraphics[width=\columnwidth]{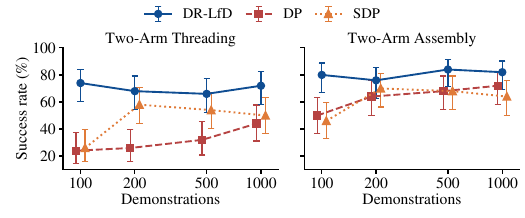}
\caption{Success rates on two DexMimicGen tasks with different numbers of demonstrations, evaluated over 50 trials per setting. Error bars show 95\% Wilson confidence intervals. }
\label{fig:dexmimicgen}
\end{figure}

\paragraph{Ablation on Demonstration Quantity}
To evaluate data efficiency, we carry out an ablation study on DexMimicGen \cite{jiang2025dexmimicgen} with varied numbers of demonstrations per task. We focus on two representative bimanual tasks: \textit{Two-Arm Threading} and \textit{Two-Arm Assembly}, where threading and insertion stages are contact-rich and precision-sensitive.
We compare \ours against end-to-end imitation policies DP \cite{chi2023diffusion} and SDP \cite{zhu2025se}, which uses \textit{SE}(3) equivariance to promote data efficiency. We vary the number of demonstrations from 100 to 1000 per task and test each setting with 50 trials. 

As shown in Fig.~\ref{fig:dexmimicgen}, \ours trained on just 100 demonstrations outperforms DP and SDP trained with 1000 demonstrations. These results show that effective visuomotor skills can be learned from small-scale data when learning focuses on contact-rich segments. 
Nonetheless, we observe that the performance of \ours saturates as the number of demonstrations increases. This suggests that, despite improvements in object-centric primitives, the visuomotor policy remains a bottleneck in fine-grained manipulation and may benefit from post-training (e.g., RL-100~\cite{lei2025rl100}).

\begin{figure}[t]
    \centering
    \includegraphics[width=\columnwidth]{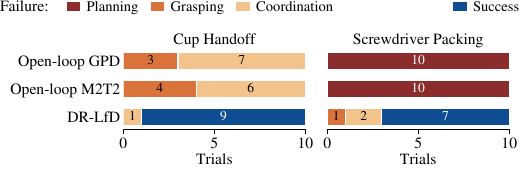}
    \caption{Real-world ablations on TAMP-inspired baselines over 10 trials per task. Planning failure, grasping failure, and subsequent coordination failure are the main failure cases.}
    \label{fig:real_generic_grasp}
\end{figure}

\paragraph{Evaluation of TAMP-inspired Contacts}
To evaluate whether traditional TAMP grasping can replace the learned primitives in \ours, we replace them with generic grasp planners or a scripted grasp on simulated and real-world tasks. In the real world, \ours (Open-loop GPD) and \ours (Open-loop M2T2) fail in all trials of \textit{Cup Handoff} and \textit{Screwdriver Packing} (Fig.~\ref{fig:real_generic_grasp}). Across all settings, these grasps fail in two ways.

First, generic grasp planners often fail to produce a feasible grasp. On LIBERO (Table~\ref{tab:libero_results}), \ours (Open-loop M2T2) has a higher success rate than \ours (Open-loop GPD), as target-only grasps from GPD \cite{ten2017grasp} are more likely to be rejected for collisions with surrounding objects. Even so, it performs poorly in cases containing flat objects, tight placements, and multiple grasps in LIBERO \textbf{Object} and \textbf{Long} set. In \textit{Screwdriver Packing}, neither baseline produces a valid grasp on the slender object.

Second, a feasible grasp can still be unsuitable for the downstream skill. In \textit{Cup Handoff}, most failures occur during bimanual coordination after a grasp is established. On DexMimicGen, a scripted top grasp leads to complete failure (0/20) in the two-arm threading task and a low success rate (7/20) in the two-arm assembly task, as the scripted contact poses do not match the demonstration distribution and result in OOD observations.

Together with the comparison against monolithic imitation policies above, these results confirm that \ours's advantage comes from combining TAMP with learned, policy-compatible primitives, addressing \textbf{Q4}.

\subsection{Real-world Robustness Stress Tests}\label{sec:stress-tests}

\begin{figure*}[t]
\centering
\includegraphics[width=\textwidth]{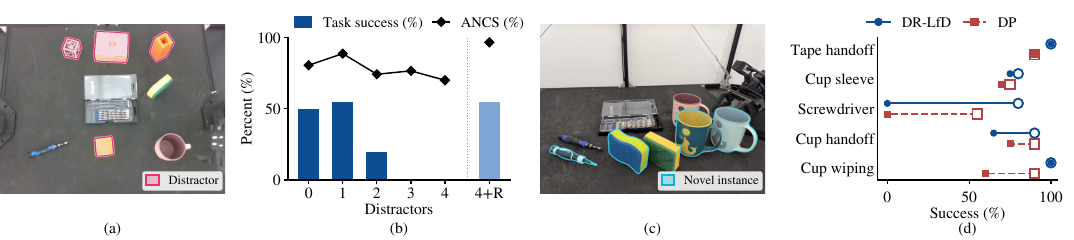}
\caption{Real-world stress tests with 20 trials per setting. (a) Clutter-test scene with four distractors. (b) \textit{Cup-Sponge-Screwdriver} task with zero to four distractors. Bars show the success rate, and diamonds show the ANCS (average number of completed steps) as a percentage of the total task steps. 4+R denotes four distractors after each skill is retrained with five additional cluttered-background demonstrations, without recollecting long-horizon demonstrations. (c) Standard and novel object instances. (d) Success rates of DP and \ours with standard (hollow markers) and novel (filled markers) instances.}
\label{fig:robustness-stress}
\end{figure*}

Fig.~\ref{fig:robustness-stress} reports three real-world stress tests in settings with adversarial object displacements, background clutter, and novel object instances.

The adversarial object displacement test is designed to evaluate the online replanning ability of \ours. Before the robot reaches a target object, an operator deliberately moves the target object away, which makes the grasp fail and triggers re-perception and local replanning, as described in Fig.~\ref{fig:overall-diagram}(d). \ours succeeds in 9, 5, and 10 of 10 trials in Object Handoff, Screwdriver Packing, and Cup Wiping, respectively, compared with 100\%, 80\%, and 100\% without displacement. The performance gap is mainly caused by imperfect re-perception or downstream skill-execution failures.

We carry out the clutter-background test on the cup-sponge-screwdriver task, whose four-distractor configuration is shown in Fig.~\ref{fig:robustness-stress}(a).
According to the results in Fig.~\ref{fig:robustness-stress}(b), the success rate falls from 50\% without distractors to 0\% with four distractors.
Most failures happen in several challenging subtasks like screwdriver packing, as indicated by the mild drop of the average number of completed steps (ANCS).  
After retraining each skill with five additional cluttered-background demonstrations (4+R in Fig.~\ref{fig:robustness-stress}(b)), the success rate rises to 55\%.
The novel-instance test is carried out in object handoff, cup-sleeve insertion, screwdriver packing, cup
handoff, and cup-wiping tasks. In each task, we replace one object with a novel instance (Fig.~\ref{fig:robustness-stress}(c)) while keeping the object category fixed. As shown in Fig.~\ref{fig:robustness-stress}(d), the visual and geometric distribution shifts affect both DP and \ours, lowering the overall task success rate. The bottleneck still lies in the visuomotor policy rollout phase, which may stall or diverge from demonstrations.
We also evaluate the contact detector on the cup, screwdriver, and sponge grasp primitives of the \textit{Cup-Sponge-Screwdriver} task under the cluttered background and novel instance scenarios. Its F1 score stays above 0.90 in all settings, and the cluttered background changes it by at most 0.01 from the original scene. It serves as a reliable skill monitor, as no real-world execution failure was caused by detector errors.
We refer readers to the project website for video recordings.

\section{Discussion} \label{sec:discussion}

\subsection{Implications for Foundation-Model-Based Robot Systems}

From the results of \ours, we draw two implications for foundation-model-based robot systems.

First, foundation-model semantics are most useful when turned into verifiable action structure. In \ours, the VLM serves as a semantic grounding layer: it extracts task-relevant objects and relations, which are transformed into contact graphs for skill abstraction and used to condition object-centric primitives.

Second, decoupled skill representations complement behavior foundation models. Our comparison with $\pi_{0.5}$ (Table~\ref{tab:real_ood}) motivates a decoupled view in which each skill is realized by the representation best suited to it---an \textit{SO}(3)-equivariant object-centric primitive, a visuomotor policy, or a planned motion. TAMP recombines these skills for new goals (Section~\ref{sec:res_combine_skills}), so a behavior foundation model could provide local visuomotor skills within this structure. In \ours, this integration improves spatial generalization, while reachability and policy-safety streams expose continuous feasibility, contact reasoning, and recovery requirements to the planner.

\subsection{Limitations Inherited from TAMP}
Despite the automated PDDL domain construction developed in Section \ref{sec:schemata}, there is still a need for manual specification in TAMP, particularly when defining customized goals (as in Section \ref{sec:res_combine_skills} ). 
It remains our future work to integrate a user-friendly interface for goal conversion, which may involve large language model translation \cite{xie2023translating}.

\begin{table}[h]
\centering
\caption{Initial planning time (s) of TAMP: mean $\pm$ std over 10 trials.}
\label{tab:scalability}
\setlength{\tabcolsep}{3pt}
\begin{tabular}{@{}L{0.28\columnwidth}|C{0.15\columnwidth}C{0.15\columnwidth}C{0.15\columnwidth}C{0.2\columnwidth}@{}}
\toprule
No.\ objects & \textbf{0} & \textbf{1} & \textbf{2} & \textbf{3} \\
\midrule
\textit{Peg-in-Hole-\texttt{unsafe} (blocking)} & $9.04 \pm 1.60$ & $35.04 \pm 12.70$ & $55.17 \pm 38.20$ & $145.51 \pm 190.00$ \\
\textit{Three-tapes (goal)}                      & ---                    & $3.16 \pm 0.74$  & $8.82 \pm 1.22$  & $36.44 \pm 10.32$ \\
\bottomrule
\end{tabular}
\end{table}

\begin{figure}[h]
\centering
\includegraphics[width=\columnwidth]{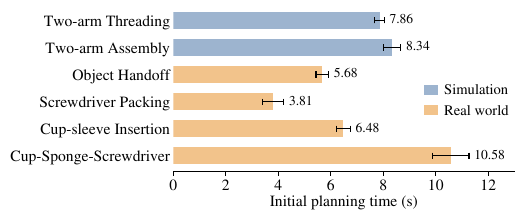}
\caption{ Initial planning time of \ours, i.e., the time to find the first feasible plan (mean $\pm$ std). The time includes TAMP search and stream computation, including inference of the equivariant networks used by learned object-centric streams.}
\label{fig:planning_time}
\end{figure}

Besides the manual effort, TAMP is also notorious for its exponentially growing complexity w.r.t the number of involved symbols, which also applies to \ours due to our usage of the \textit{focused} solver \cite{garrett2018sampling}.
We conducted experiments to empirically analyze the scalability of \ours w.r.t increasing number of objects. 
Specifically, experiments were conducted under two variations of conditions: 
In the simulated \textit{Peg-in-Hole-\texttt{unsafe}} scenario (see Fig. \ref{fig:unsafe-unreach} of Section \ref{sec:constraint_handle}), we increase the number of \textbf{\textit{blocking}} objects that block the peg-in-hole operation, i.e., the plunger-like obstacles rest on the table center; 
In the \textit{Three-tapes} scenario (see Fig. \ref{fig:long-tasks}  of Section \ref{sec:res_combine_skills}), we increase the number of \textbf{\textit{goal}} objects, i.e.,  the tapes to be handed off and placed into the basket. 
Table \ref{tab:scalability} displays the averaged initial planning time (in seconds) with respect to the number of objects, calculated across 10 runs.
The reported standard deviations arise mainly from the randomized continuous samplers used in stream evaluation, i.e., IK solving, sampling-based motion planning, and generic grasp planning. As each obstacle relocation must jointly certify a grasp, a collision-free configuration, and a collision-free trajectory under tight workspace collisions, it may take indefinitely long to find a valid solution.
We observe an exponential growth in initial planning time with respect to the complexity of the constraints and goals, which aligns with the PSPACE-hard complexity in most modern TAMP methods \cite{garrett2020pddlstream,garrett2021integrated}.
Nevertheless, when subgoals are specified in the goal description, as in the tasks reported in Fig.~\ref{fig:planning_time}, the initial planning time stays below 11\,s for all tasks, so the planning cost remains practical. 
Due to the modular design of \ours, TAMP computation can potentially be accelerated by more advanced solvers such as cuTAMP \cite{shen2024differentiable}. 

\subsection{Limitations Inherited from  Equivariant Learning}

Similar to common practice in TAMP \cite{garrett2020online,pan2024task,curtis2022long}, \ours uses depth measurements from a calibrated depth sensor on demand. Access to 3D object observations is required for learning object-centric primitives. 
Although \ours exhibits greater robustness to OOD object distributions, the incomplete or noisy depth data may degrade the quality of \textit{SO3}-equivariance learning, as seen in the \texttt{Leaky} scenarios (Section~\ref{sec:real-exp}). 
This issue can be alleviated through improved engineering, such as using a learned depth estimator \cite{wen2025foundationstereo} for higher-fidelity depth or applying point cloud completion to obtain more complete object models \cite{liu2020morphing}. 

The reliance on metric depth images additionally imposes stricter sensor requirements in comparison with those image-only visuomotor policies.  
Although there is increasing interest in learning equivariant representations directly from RGB (e.g., \cite{hu20253d,wang2025practical}), their performance remains limited and their applicability is narrow, as discussed in Section \ref{sec:relatedwork}. To ensure reliability, \ours confines equivariant learning to object-centric applications, a conservative but practical design choice. Exploring equivariant learning without depth measurements is an open direction for future work.

\section{Conclusion} \label{sec:conclusion}
We proposed \ours, a data-efficient system that leverages the advantages of both visuomotor policies and classical planning for generalized and long-horizon robotic manipulation tasks. The philosophy of \ours is to decompose and learn atomic skills from demonstrations, and reorganize learned skills to complete tasks with potentially novel goals and setups while complying with physical constraints. Specifically, in the decomposition phase, skills are partitioned and categorized by a VLM-assisted contact-aware approach according to changes in contact-relationships. Each skill is learned as a visuomotor policy or an object-centric primitive, and is converted to a TAMP-compatible form. In the reorganization phase, \ours automatically constructs the TAMP problem, with physical constraints carefully implemented in test streams. The online planning and replanning scheme further enhances the reliability of collision-free motion planning.
Based on results across diverse tasks and settings, \ours shows resilience to random initial conditions that are unseen in training data and even unattainable through teleoperation alone. By systematically integrating learned skills into a TAMP framework,  \ours demonstrates its ability to combine previously acquired skills to solve long-horizon tasks while complying with novel hard constraints. Nevertheless, the limitations inherited from its building blocks remain critical hurdles for better performance and easier deployment. It is an important future direction to study how to circumvent those limitations while preserving the merits of planning and learning methods.

\bibliographystyle{IEEEtran}   %
\bibliography{IEEEabrv,bib/reference} 

\appendices

\appendices
\section{Spatial Generalization and Capability Boundary}
\label{sec:appendix-spatial-boundary}

For objects movable by a skill in the learned repertoire, the learned objective-centric primitives and keyposes restore in-distribution observation for subsequent policy. Let $s = \{s_h, s_o, s_{h-o}\}$ denote the robot, object, and robot--object contact states. Assuming each sensing modality depends only on the relevant subset of the state, the observation model factorizes as
\vspace{-0.05in}
\begin{equation}
P(\mathcal{O} \mid s)
= P(q \mid s_h)\, P(I_\mathrm{wrist} \mid s_{h-o})\, P(I_{3rd} \mid s_h, s_o),
\end{equation}
where $q$, $I_\mathrm{wrist}$, and $I_{3rd}$ denote proprioception, wrist-camera images, and third-view images, respectively.

Planning to predicted keyposes provides in-distribution proprioception $q$. Learned primitives (Section~\ref{sec:primitivelearn}) provide in-distribution contact states $s_{h-o}$ and thus wrist observations $I_\mathrm{wrist}$. For rigidly attached objects, these contact states and robot keyposes also determine the object states $s_o$, supporting in-distribution third-view observations $I_{3rd}$. Together, $\Phi_\tau$ and $\Phi_{\mathrm{kp}}$ prepare the observations required to initiate the visuomotor policy $\pi$.

\section{Motion Primitives Description} \label{sec:planned_actions}

In \ours, we currently include several  object-centric motion primitives:
$$ \{ \pddl{Transit}, \pddl{Transfer}, \pddl{Pick}, \pddl{Place}, \pddl{Drop} \}.$$
\texttt{Transit} and \texttt{Transfer} move the arm with an empty gripper or while grasping an object, respectively; \texttt{Pick} grasps an object, \texttt{Place} releases it onto a surface, and \texttt{Drop} releases it into a container. As these are standard prehensile and free-space primitives, we hand-craft their schemata following the standard PDDLStream definitions of classical TAMP systems~\cite{garrett2018sampling, vu2024coast}. The explanation of the predicates and associated variables used throughout \ours is summarized in Table~\ref{tab:predicates}.

\begin{table}[ht]
\centering
\caption{Explanation of Predicates.}
\label{tab:predicates}
\begin{tabular}{@{}L{0.38\columnwidth}L{0.54\columnwidth}@{}}
\toprule
\textbf{Predicate Name} & \textbf{Explanation}\\ \midrule
$\pddl{AtConf}(h, q)$  &  The joint angles of  robot arm $h$ are $q$\\
$\pddl{AtGrasp}(h, o, g)$  &  The object $o$ is grasped by the robot arm $h$ with a grasp pose $g$                 \\   
 $\pddl{AtPose}(o, p)$ & The object $o$ is at pose $p$    \\  
  $\pddl{On}(o_2, o_1 )$ &  The object $o_2$  is placed on  object $o_1$                \\  
    $\pddl{In}(o_2,o_1)$    & The object $o_2$ is inside a container $o_1$          \\
      $\pddl{HandEmpty}(h)$    & The gripper of the robot arm $h$ is free          \\ 
    $\pddl{Surface}(o)$    & The object $o$ can be a supporting surface         \\ 
      $\pddl{Movable}(o)$    & The object $o$ can be grasped          \\
      $\pddl{Container}(o)$    & The object $o$ is a container  \\
  \bottomrule
\end{tabular}
\end{table}

{\section{Failure Cases in Grounding Skills and Objects}\label{sec:appendix-grounding-failures}}

This section discusses the details of grounding. In practice, we assumed step-by-step textual guidance without ambiguity. Such descriptions are available for datasets like LIBERO \cite{liu2023libero} but are generally unavailable in real-world tasks or DexMimicGen \cite{jiang2025dexmimicgen}. Therefore, we manually provide these descriptions to help the VLM ground skills and objects.

Open-set object segmentation introduces another failure mode: limited vocabulary. We observed no such segmentation failures with GroundedSAM \cite{ren2024grounded} in our quantitative real-world experiments, which use common household items. However, the plunger-like obstacle in Section~\ref{sec:constraint_handle} (Fig.~\ref{fig:constraint-visualize}) is unrecognized in roughly half of the trials. This could be addressed by a more advanced open-world segmentation tool with broader vocabulary.

To quantify the reliability and human-checking cost of this grounding stage, we further evaluate the VLM output against ground-truth scene-graph sequences with no human intervention. As shown in Table~\ref{tab:human_VLM}, we run the VLM on 20 episodes per task and report three metrics: SGCorr, which measures per-skill cumulative scene-graph correctness; Edge-Op F1, which measures the F1 score over the full multiset of predicted contact edge operations; and MaxEdit, which measures the maximum per-skill number of edge-operation edits required to match the ground truth. We also report the human check time needed to inspect and correct the structured output before downstream skill segmentation.

\begin{table}[t]
\centering
\caption{Evaluation of VLM-assisted skill abstraction ($N=20$ runs per task, no human intervention) and the  human check time (five annotators). }
\setlength{\tabcolsep}{1pt}
\begin{tabular}{@{}L{0.32\columnwidth}C{0.14\columnwidth}C{0.16\columnwidth}C{0.14\columnwidth}C{0.18\columnwidth}@{}}
\toprule
\textbf{Task} & \textbf{SGCorr}$\uparrow$ & \makecell[c]{\textbf{Edge-Op}\\\textbf{F1}$\uparrow$} & \textbf{MaxEdit}$\downarrow$ & \makecell[c]{\textbf{Human}\\\textbf{Check (s)}$\downarrow$} \\
\midrule
Object Handoff       & 0.93 & 0.98 & 0.10 & 21.00 \\
Screwdriver Packing  & 0.50 & 0.76 & 2.60 & 44.10 \\
Cup-sleeve Insertion & 1.00 & 1.00 & 0.00 & 13.30 \\
Threading            & 1.00 & 1.00 & 0.00 & 14.80 \\
Assembly             & 0.58 & 0.92 & 1.25 & 26.50 \\
LIBERO-spatial       & 1.00 & 1.00 & 0.04 & 15.30 \\
LIBERO-object        & 1.00 & 1.00 & 0.00 & 10.10 \\
LIBERO-long(1,5,8)   & 0.53 & 1.00 & 1.33 & 59.70 \\
\bottomrule
\end{tabular}
\label{tab:human_VLM}
\vspace{-0.2in}
\end{table}

These results support three observations. \textbf{High Native Correctness:} the VLM must first commit to the ordered skill sequence and then attach each edge operation to a specific skill, leading to high correctness on most tasks. \textbf{Low Downstream Impact:}  bimanual-internal edge errors do not affect downstream learning, because complex bimanual skills are treated as black-box atomic skills. \textbf{Effortless Human Validation:} the main correction-relevant failure mode is instance ambiguity(e.g., two moka pots in \textit{LIBERO-long}), but it remains below one minute for manual checking.

\end{document}